\RequirePackage[hyphens]{url}
\documentclass{article}
\usepackage{microtype}
\usepackage{graphicx}
\usepackage{subfigure}
\usepackage{booktabs} 
\usepackage{makecell}
\usepackage{multicol} 
\usepackage{hyperref}
\usepackage{xspace}
\usepackage[inline]{enumitem}
\usepackage{balance}

\newenvironment{denseitemize}{
\begin{itemize}[topsep=2pt, partopsep=0pt, leftmargin=1.5em]
  \setlength{\itemsep}{2pt}
  \setlength{\parskip}{0pt}
  \setlength{\parsep}{0pt}
}{\end{itemize}}

\newenvironment{denseenum}{
\begin{enumerate}[topsep=2pt, partopsep=0pt, leftmargin=1.5em]
  \setlength{\itemsep}{2pt}
  \setlength{\parskip}{0pt}
  \setlength{\parsep}{0pt}
}{\end{enumerate}}

\def\ie{{i.e. }}
\def\eg{{e.g. }}
\def\etal{{et al.\xspace}}

\newcommand{\unaryminus}{\scalebox{0.75}[1.0]{\( - \)}}

\usepackage[accepted]{mlsys2020}

\mlsystitlerunning{Characterizing and Taming Model Instability Across Edge Devices}

\begin{document}

\twocolumn[
\mlsystitle{Characterizing and Taming Model Instability Across Edge Devices}



\mlsyssetsymbol{equal}{*}

\begin{mlsysauthorlist}
\mlsysauthor{Eyal Cidon}{equal,st}
\mlsysauthor{Evgenya Pergament}{equal,st}
\mlsysauthor{Zain Asgar}{st}
\mlsysauthor{Asaf Cidon}{col}
\mlsysauthor{Sachin Katti}{st}
\end{mlsysauthorlist}

\mlsysaffiliation{st}{The Department of Electrical Engineering, Stanford University, Stanford, California, USA}
\mlsysaffiliation{col}{Electrical Engineering Department, Columbia University, New York, New York, USA}

\mlsyscorrespondingauthor{Eyal Cidon}{ecidon@stanford.edu}
\mlsyscorrespondingauthor{Evgenya Pergament}{evgenyap@stanford.edu }

\mlsyskeywords{Machine Learning, MLSys, Mobile Phones}

\vskip 0.3in
\begin{abstract}
The same machine learning model running on different edge devices may produce highly-divergent outputs on a nearly-identical input. Possible reasons for the divergence include differences in the device sensors, the device's signal processing hardware and software, and its operating system and processors. This paper presents the first methodical characterization of the variations in model prediction across real-world mobile devices. We demonstrate that accuracy is not a useful metric to characterize prediction divergence, and introduce a new metric, instability, which captures this variation. We characterize different sources for instability, and show that differences in compression formats and image signal processing account for significant instability in object classification models. Notably, in our experiments, 14-17\% of images produced divergent classifications across one or more phone models.
We evaluate three different techniques for reducing instability. In particular, we adapt prior work on making models robust to noise in order to fine-tune models to be robust to variations across edge devices. We demonstrate our fine-tuning techniques reduce instability by 75\%.
    
\end{abstract}
]



\printAffiliationsAndNotice{\mlsysEqualContribution} 

\section{Introduction}


Machine learning (ML) models are increasingly being deployed on a vast array of devices, including a wide variety of computers, servers, phones, cameras and other embedded devices~\cite{FB-ML-inference-on-edge}. The increase in the diversity of edge devices, and their bounded computational and memory capabilities, has led to extensive research on optimizing ML models for real-time inference on the edge.
However, prior work primarily focuses on the model's properties rather than on how each device introduces variations to the input. They tend to evaluate the models on public, well-known and hand-labeled datasets. However, these datasets do not necessarily represent the transformations performed by these devices on the input to the model~\cite{DBLP:conf/icml/RechtRSS19,biased-datasets}. The wide variety of sensors and hardware, different geographic locations \cite{phones-different-countries} and processing pipelines all effect the input data and the efficacy of the ML model. 

\begin{figure}[t!]
\includegraphics[width=0.15\textwidth]{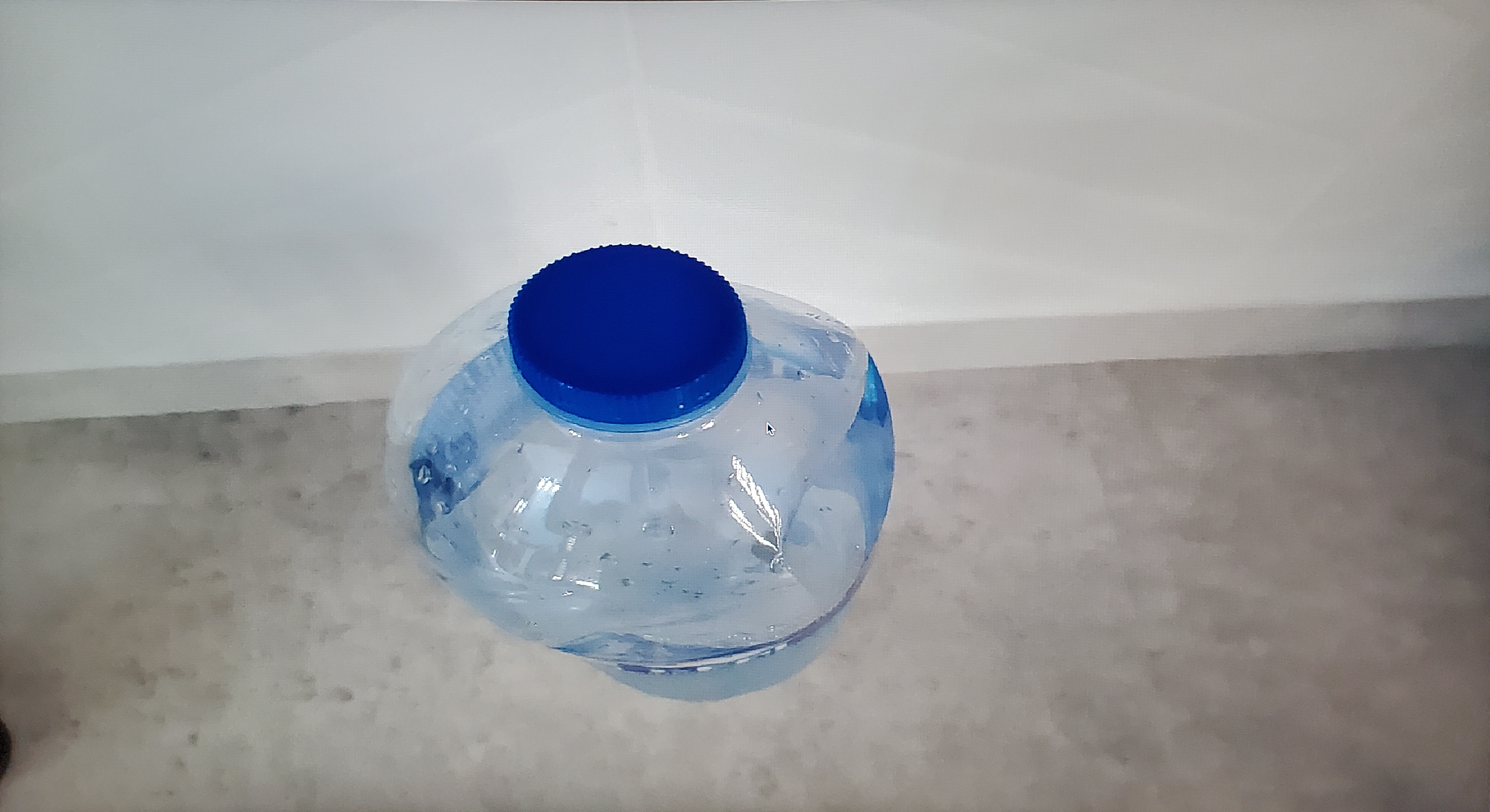} 
\includegraphics[width=0.15\textwidth]{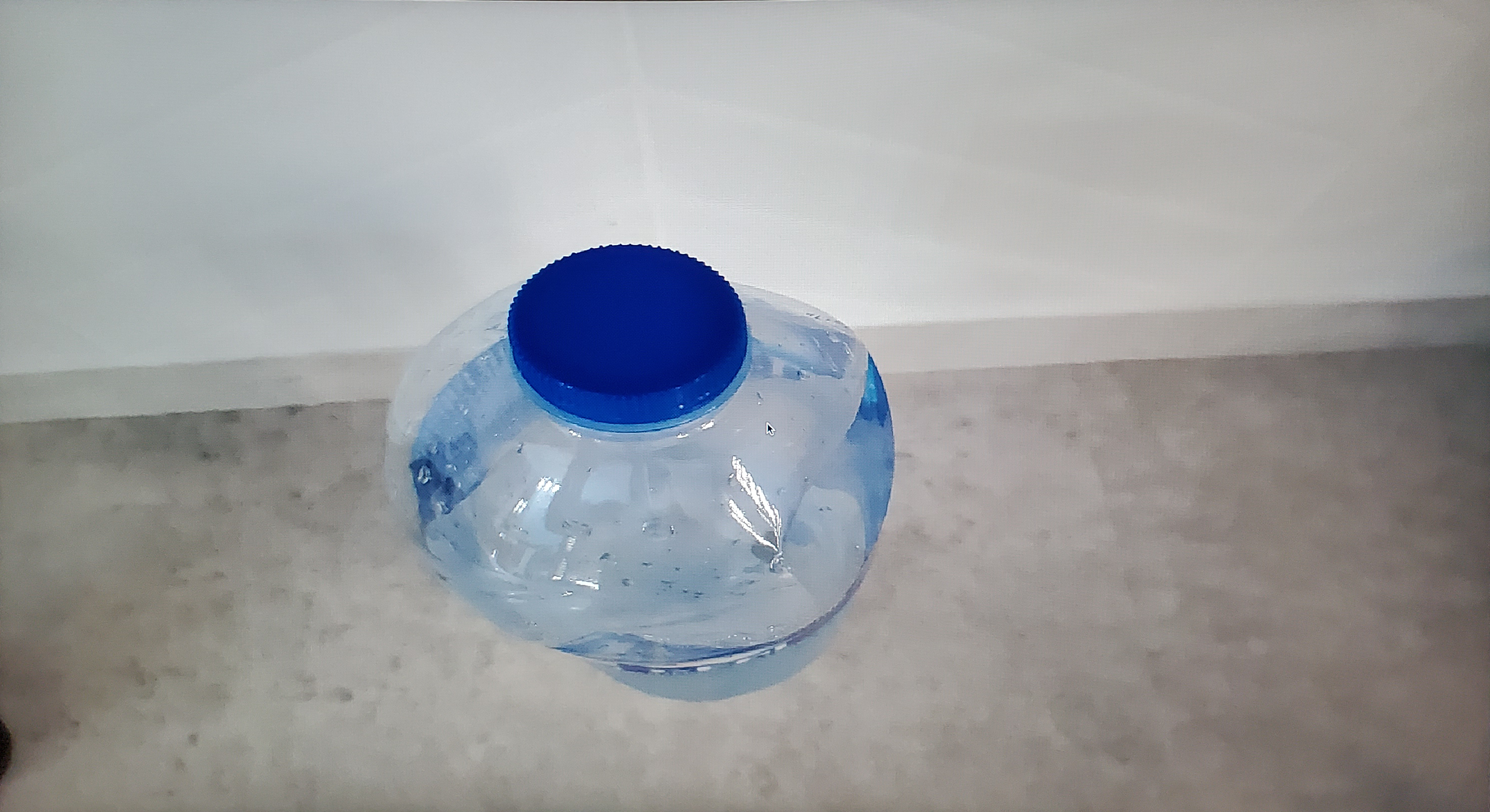}
\includegraphics[width=0.15\textwidth]{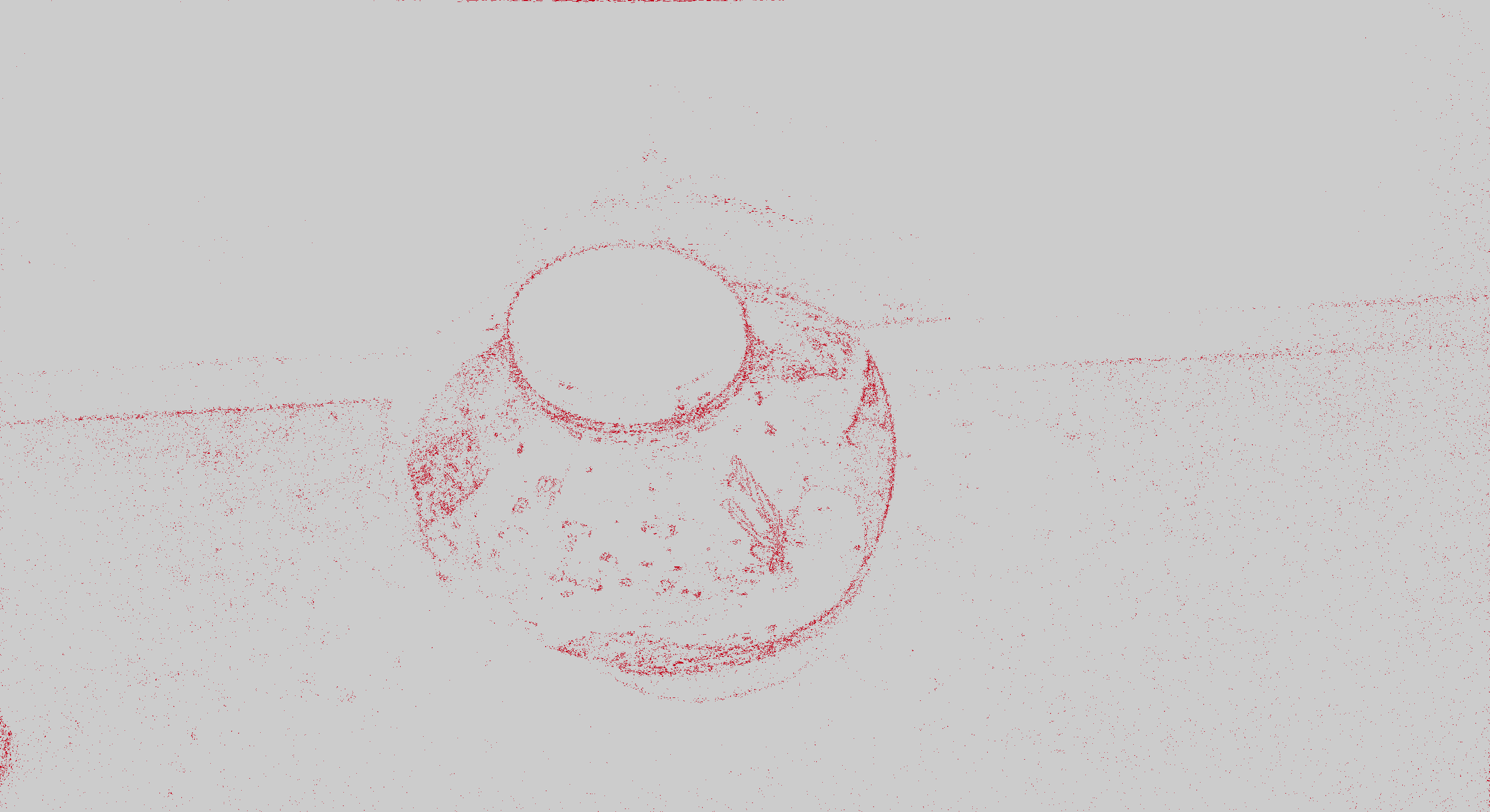}
\vspace{-0.25cm}
\caption{Two photos taken one second apart with Samsung Galaxy S10 in a controlled environment, where the phone was not touched between the two shots. MobileNetV2 assigns the incorrect label ``bubble'' on the left image and the correct label ``water bottle'' on the middle image. The right image shows the pixel difference between both photos. The red dots represent the pixels within a range higher than 5\%. There is a very small pixel difference between the images, yet it is sufficient to cause instability.}
\label{fig:burst}
\end{figure}

To demonstrate how even small real-world variations to the input can affect the output, in Figure~\ref{fig:burst} we show an example where pictures of the same object taken by the same phone, one second apart, from a fixed position (without moving the phone), produce different classification results on the same model. Such divergence occurs even more frequently when the same model is run on different devices. In order to understand how the model will perform when run on many different devices, it is important to ensure consistent model accuracy, and create representative training and evaluation datasets that will be robust to variations across devices and sensors.


To capture this variability, we introduce the notion of \emph{instability}, which denotes the probability that the same model outputs different contradictory predictions on nearly identical input when run on different edge devices.
Running models on a wide array of edge devices can create many different opportunities for instability. Some of the instability is caused by changes to the inputs to the model, for example by the usage of different sensors on different devices (\eg camera lenses), applying different transformations on the raw captured data (\eg image processing pipelines), and saving the data in different formats (\eg compression). Further instability may be introduced by hardware on the device (\eg GPUs handling of floating points) or the operating system (\eg the OS stores pictures in a particular format).

Prior work on model robustness has either focused on simulating the sources of instability~\cite{buckler2017reconfiguring}, or on making models robust to adversaries~\cite{kurakin2016adversarial, bastani2016measuring, cisse2017parseval}. While adversarial learning is important for model robustness, the vast majority of instability in many applications is not caused by adversaries, but rather by normal variations in the device's hardware and software. Therefore, we lack a solid understanding of the \emph{sources of instability in real-world edge devices}.

In this work, we conduct the first systematic characterization of the causes and degree of variance in convolutional neural network (CNN) computer vision model efficacy across existing mobile devices. We conduct four sets of experiments, in which we try to isolate different sources of variance, and evaluate the degree of instability they cause. In each experiment below, the same model is used while we vary the input or the operating conditions and the outputs are compared:
\begin{denseenum}
\item \textbf{End-to-end instability across phones:} We evaluate end-to-end instability across 5 mobile devices in a lab environment, in which the phones' position and lighting conditions are tightly controlled.
\item \textbf{Image compression algorithm:} We evaluate the effect of image compression techniques, by compressing a raw image using different algorithms.
\item \textbf{Image signal processor (ISP):} We estimate the effect of using different ISPs, by comparing raw converted images using ImageMagick and Adobe Photoshop.
\item \textbf{Processor and operating system:} We evaluate the effect of the device's OS and hardware by running inference on the same set of input images across 5 phones.
\end{denseenum}

Our characterization leaves us with several takeaways. First, we demonstrate that accuracy fails to account for the lack of consistency of predictions across different devices, and motivate the need to focus on minimizing instability as an objective function. Second, we show that instability across devices is significant; between 14-17\% of images classified by MobileNetV2 have divergent predictions (both correct and incorrect) in at least two phones. Third, we show that a significant source of instability is due to variable image compression and ISPs. Finally, we do not find evidence that the devices' processors or operating system are a significant source of instability.

While the focus of this work is on systematically characterizing the sources of instability, we also provide a preliminary analysis of mitigation techniques. First, we explore whether fine-tuning the training process by augmenting the training dataset with noise or with photos taken from multiple different phone models would make the model more robust. Inspired by prior work on noise robustness~\cite{StabilityTraining}, we show that augmenting the training data set with such ``noise'', can reduce the instability by up to about 75\%.  Second, once a model is trained, instability can be further reduced if the model can either operate on raw images (when feasible) or by allowing the classification to display additional results (\eg the top three results instead of the top one result). Our future work will focus on a more systematic development of instability mitigation techniques to handle edge variance. 

The paper makes the following contributions:
\begin{denseitemize}
\item{\textbf{Instability:}} We motivate and propose instability, a new metric for evaluating the variability of model predictions across different edge devices.
\item{\textbf{Characterization:}} We present the first comprehensive characterization of the end-to-end degree and potential sources of instability on real-world mobile devices.
\item{\textbf{Mitigation:}} We propose and evaluate three potential mitigation strategies for instability.
\item{\textbf{Stability Training for Devices:}} We adapt a prior approach to make computer vision models robust to random noise~\cite{StabilityTraining} for reducing instability across devices.
\end{denseitemize}
\section{Background}

In this section, we provide background for why inference is increasingly being conducted on edge devices, and introduce and define the model instability problem.

\subsection{Inference on Edge Devices}

While much of model training occurs on servers or datacenters, there is an increasing trend of pushing inference to edge devices~\cite{FB-ML-inference-on-edge,lee2019device}. The advantages of conducting inference on edge devices over a centralized location are manifold. First, it removes the need to transfer the input data over the network. Second, it can provide lower latency. Third, users may not be always be in settings where they have a stable connection. Fourth, it can ensure the input data collected on the edge remains private.

There are many examples of real-time applications that conduct inference on the edge, including augmenting images taken from a camera~\cite{FB-ML-inference-on-edge,ignatov2017dslr}, localizing and classifying objects~\cite{speed-accuracy}, monitoring vital health signals~\cite{health-monitoring} and recognizing speech for transcription or translation~\cite{speech-recog-edge}.

However, running inference on different edge devices, each with its own hardware and software, as well as different sensors with which they capture input data, creates very heterogeneous environments. These different environments lead the model to behave differently, even on seemingly identical inputs.

\subsection{Model Instability}

While the traditional ML metrics, of accuracy, precision and recall are extremely important, they do not capture how well a model performs \emph{across environments}. For this purpose, we define a new metric, called \textbf{model instability}, which denotes when a model conducts inference in different environments, and returns significantly different results on near-identical inputs~\cite{StabilityTraining}.

Photography always introduces some random noise during image acquisition from the sensor~\cite{boncelet2009image}. Due to this, models have some degree of instability, even when they are run on exactly the same edge device. To illustrate this, we run the following simple experiment. Figure~\ref{fig:burst} depicts two photos of a water bottle from the same Samsung Galaxy S10. The photos were taken couple of seconds apart using Android debug bridge, without touching the phone or changing its location. Both images seem identical to the naked eye. However, when we run MobileNetV2~\cite{mobilenetv2} on both models, the one on the left returns the ``bubble'' class (incorrect), while the model predicts ``water bottle'' class (correct) for the center image. The image on the right shows the pixels where the difference between in the images is higher than 5\%. This experiment shows that even on the same phone, two photos taken within a very short timespan may lead to different predictions.

In order to quantify the measure of instability of a model, we define a prediction as \emph{unstable} if in at least one environment it returns a correct class, and in at least one other environment it returns a clearly incorrect class. We do not compare the predictions between different environments in the case where all the predictions are incorrect, because it is difficult to say whether a particular classification is more ``incorrect'' than another. 

Much of prior work focused on making models robust to adversaries~\cite{kurakin2016adversarial, bastani2016measuring, cisse2017parseval}. In contrast, our focus in this work is on making models robust to naturally occurring variations in model output due to different processors, software and sensors.

\section{Experiment Overview}

This section details our data collection, experimental setup and the goals of our experiments.

\subsection{Data Collection}
\label{sec:data-collection}

In order to run our experiments we collected images representing a subset of 5 classes from ImageNet~\cite{imagenet}: water bottle, beer bottle, wine bottle, purse and backpack. The images were a mix of images scraped from Flicker, images downloaded from Amazon and Amazon Prime Now and photos we took ourselves. We collected a total of 1,537 images.

\subsection{Experimental Setup}

\begin{figure}[t!]
\centering     
\subfigure[end-to-end experiment]{\includegraphics[width=0.47\textwidth, height=3.5cm]{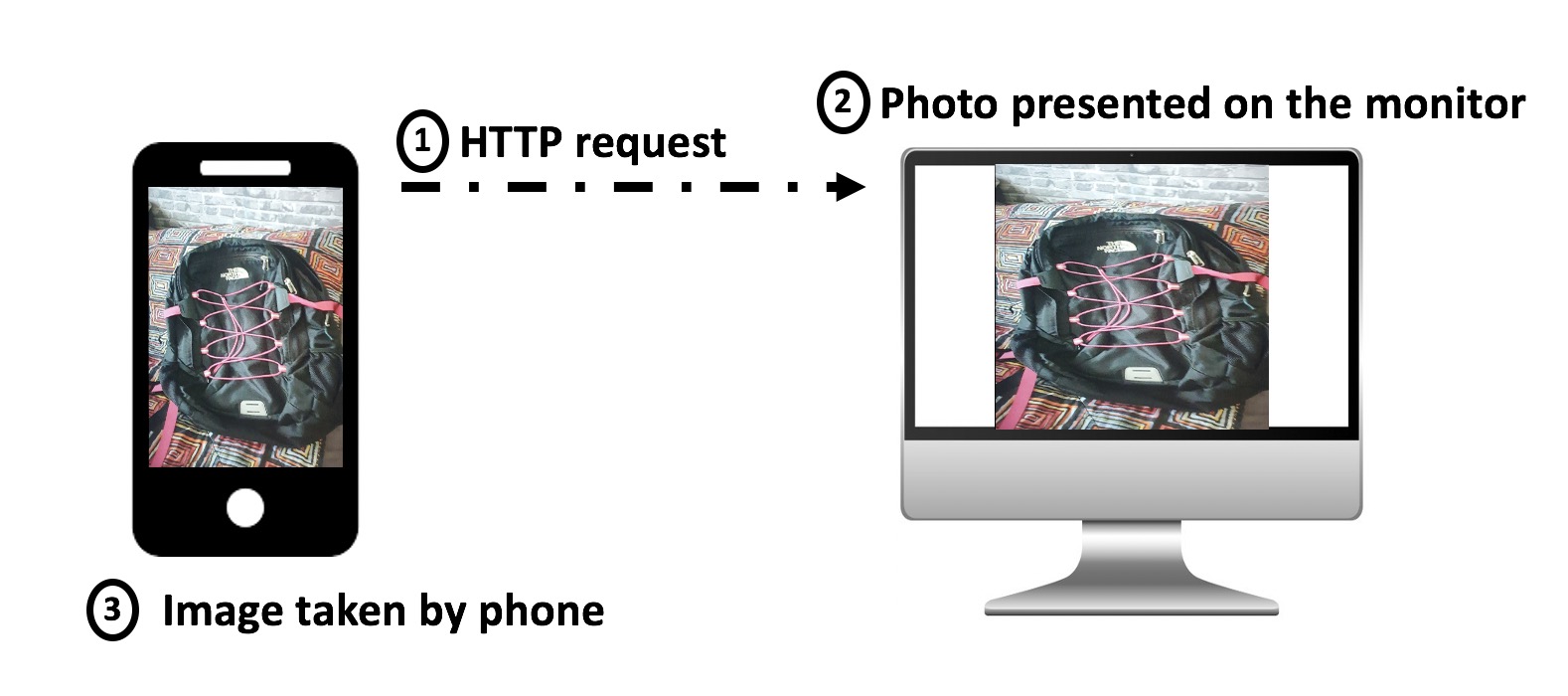}}
\subfigure[iPhone XR]{\includegraphics[width=0.15\textwidth, height=2.1cm]{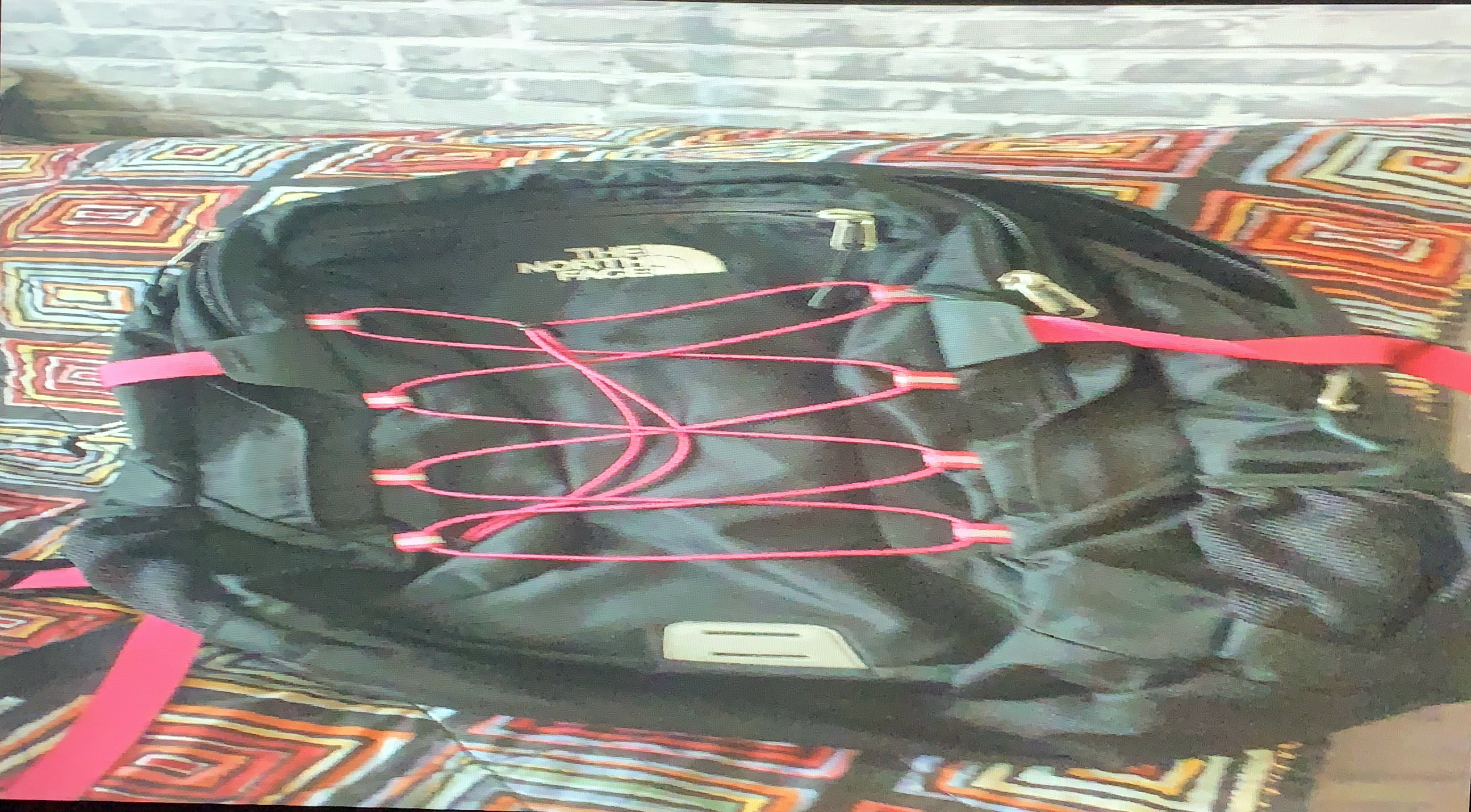}} \hspace{10mm}
\subfigure[Galaxy S10]{\includegraphics[width=0.15\textwidth, height=2.1cm]{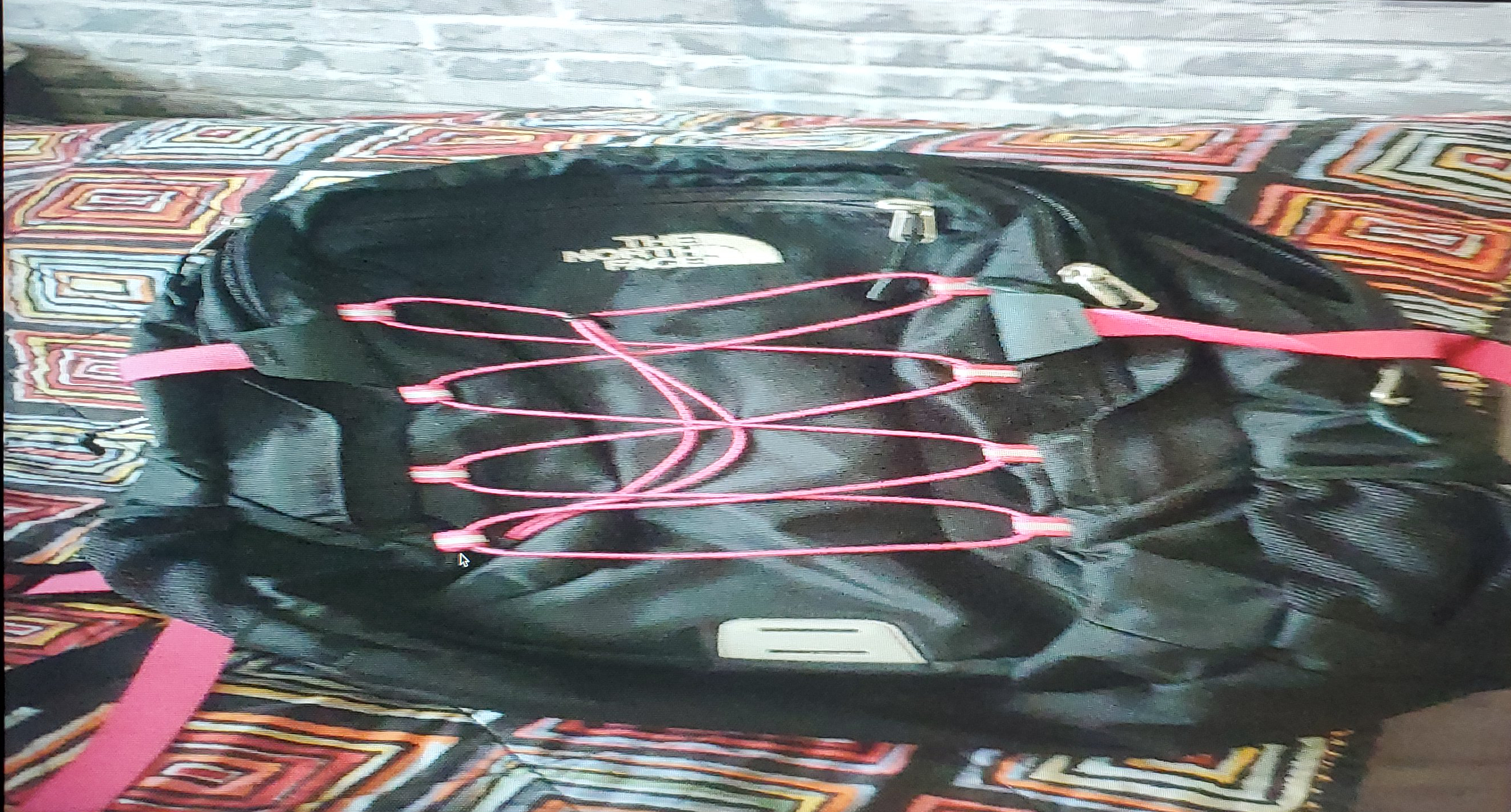}}
\vspace{-2.8pt}
\caption{(a) Illustration of experimental setup. We wrote two open-source applications: one for Android and one for iOS. Each phone sends an HTTP request to a server to display an image. After the image is presented on the monitor, the photo is taken by the phone. 
(b) Example of an iPhone photo incorrectly classified as ``pillow''. (c) Example of Samsung Galaxy S10 photo correctly classified as ``backpack''.\label{fig:system-design}}
\end{figure}

Our goal is to eliminate as much as possible any source of variability that is external to internal characteristics of the phone (\eg lighting, position, the object being photographed).
To this end, we designed the following controlled experimental setup. We placed 5 phones on a camera mount in front of a computer screen in a closed room with light-blocking curtains.
The phones took a series of photos of objects presented on the computer screen from the same angle. The objects and photos projected on the screen where taken from our collected dataset described previously (\S\ref{sec:data-collection}). Each phone was presented identical photos on the screen.
\begin{table}[h]
\caption{Phones models used in experiments.\label{tab:phone-details}}
\vskip 0.15in
\begin{center}
\begin{small}
\begin{sc}
\begin{tabular}{ll}
\toprule
Phone & Model \\
\midrule
Samsung Galaxy S10 & SM-G973U1 \\
LG K10 LTE & K425 \\
HTC Desire 10 Lifestyle & DESIRE 10 \\
Motorola Moto G5 & XT1670 \\
iPhone XR & A1984 \\
\bottomrule
\end{tabular}
\end{sc}
\end{small}
\end{center}
\vskip -0.1in
\end{table}
To ensure the phone and computer screen are synchronized, the time at which the photos are taken from the phone is controlled by apps, which we wrote for both iOS and Android\footnote{We will release the code for both apps on Github.}. Figure~\ref{fig:system-design} depicts the process: the app communicated with the computer screen, and determines which photo is currently displayed on the screen. The screen displays the photo, and then the app takes a picture of the screen from the device.
We repeated this sequence on each of the distinct objects, on 5 different angles (left, center-left, center, center-right and right) with fixed heights. In total, we took 68,125 photos. 

Throughout the experiments presented in the paper, we evaluate the performance of a single MobileNetV2~\cite{mobilenetv2} model with fixed weights, on the photos taken by the different phones. The model weights were pre-trained on ImageNet.
The list of the phones used in our experiments are available in Table~\ref{tab:phone-details}.

To evaluate the results, we verified whether images were classified correctly or not by hand. For some predictions, there can be more than one possible correct label. For instance, ``wine bottle'' and ``red wine'' overlap in ImageNet, so for a bottle of red wine we accepted both "wine bottle" and "red wine". If an image contained more than one object, we only considered the object that is clearly in the foreground of the image.

\subsection{Goals}

We run four sets of experiments, presented in the next sections: (a) end-to-end experiments, whose goal is to measure the end-to-end instability of models using the same mobile devices and across devices (\S\ref{sec:end-to-end}); (b) measuring the effect of image compression (\S\ref{sec:compression}); (c) measuring the effect of different ISPs (\S\ref{sec:ISP}); and (d) measuring the effect of the device's operating system and processor (\S\ref{sec:OS}).
\section{End-to-end Instability}
\label{sec:end-to-end}
\begin{figure}[t]
\centering
\subfigure[Accuracy by phone model.]{\label{fig:old-experiment} \includegraphics[width=0.47\columnwidth]{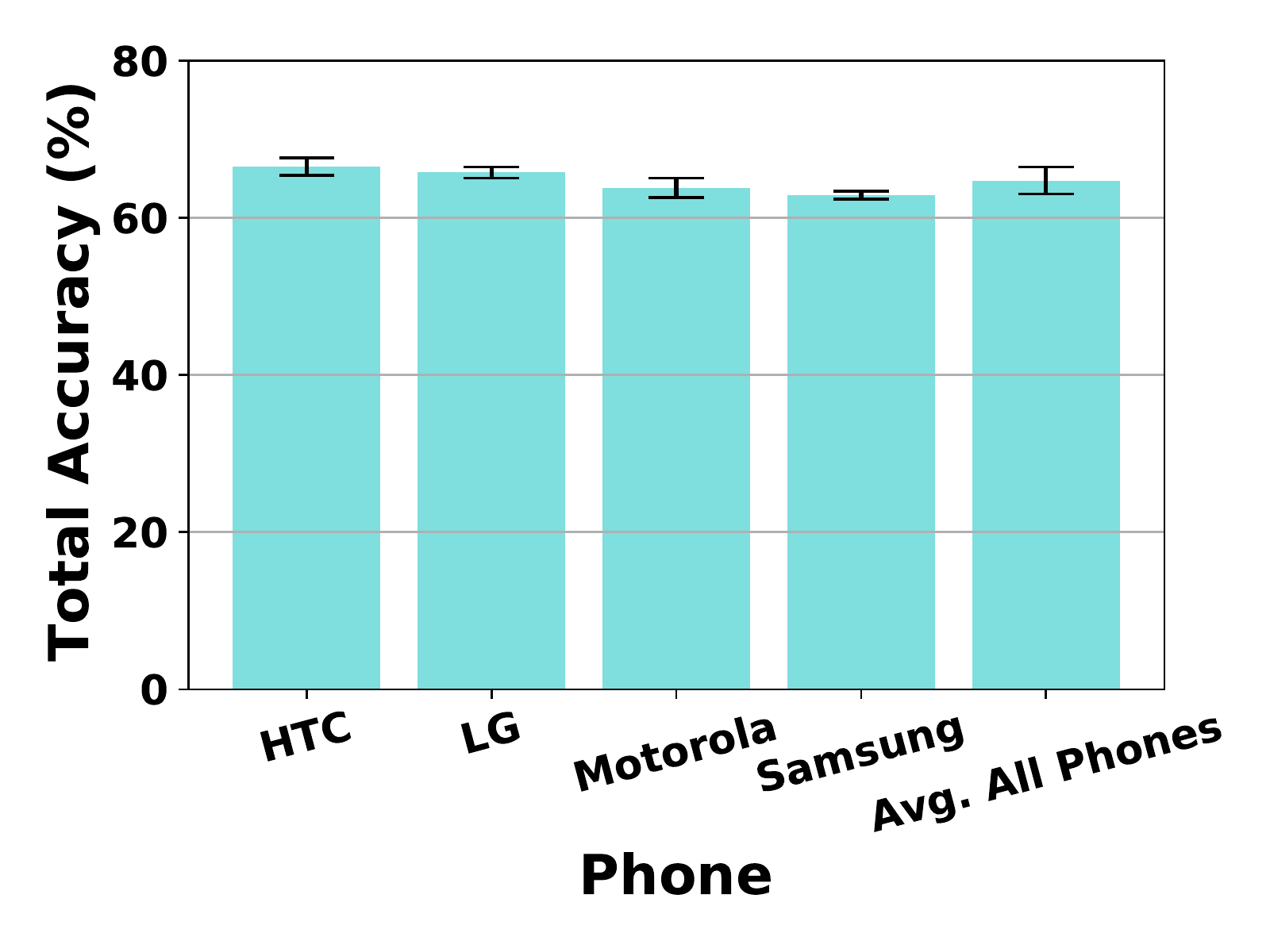}}\hfill
\centering
\subfigure[Instability by class.]{\label{fig:old-experiment-instability}
\includegraphics[width=0.47\columnwidth]{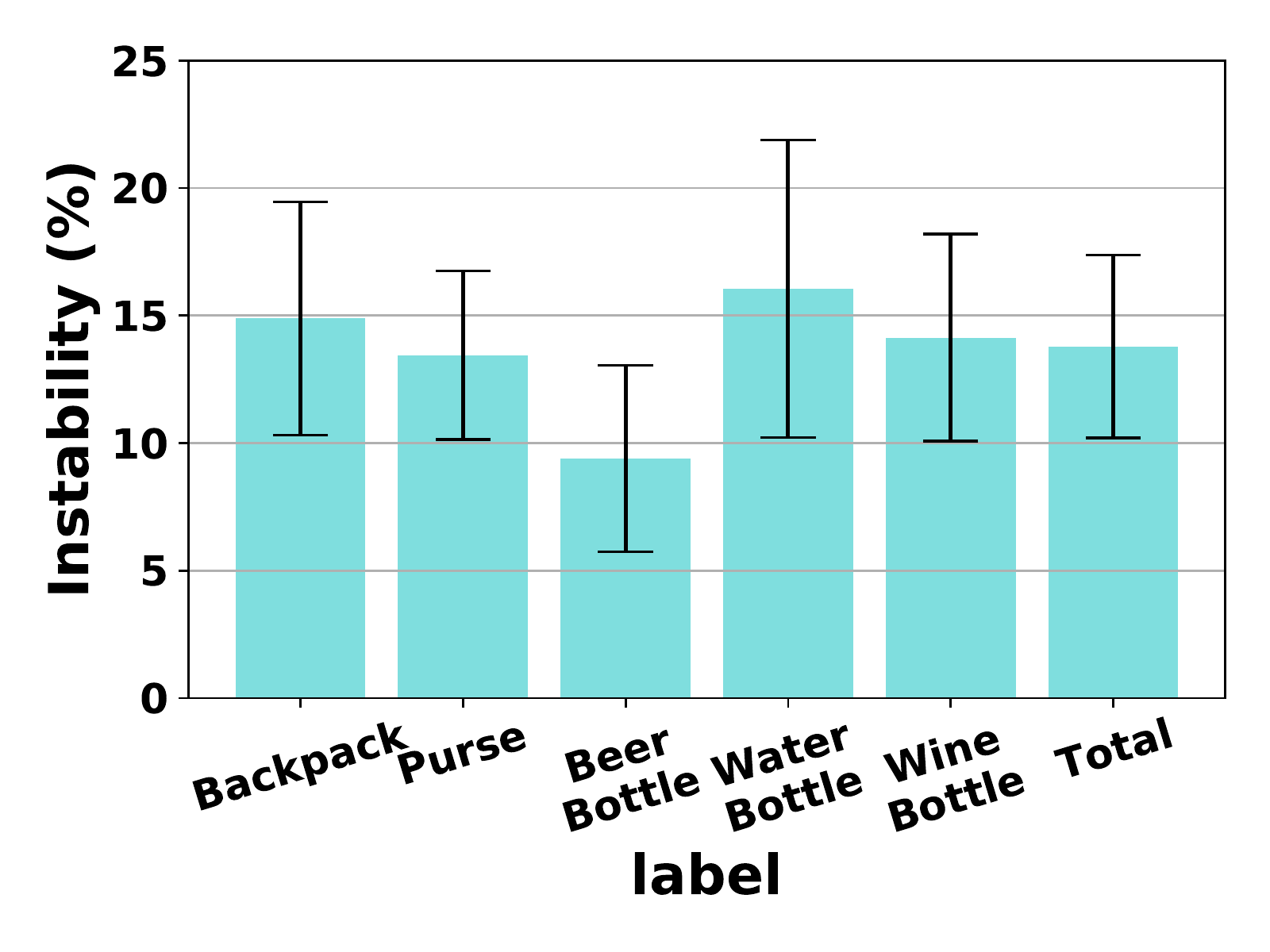}}
\centering
\subfigure[Instability by experiment angle. ]{\label{fig:old-experiment-variation}
\includegraphics[width=0.47\columnwidth]{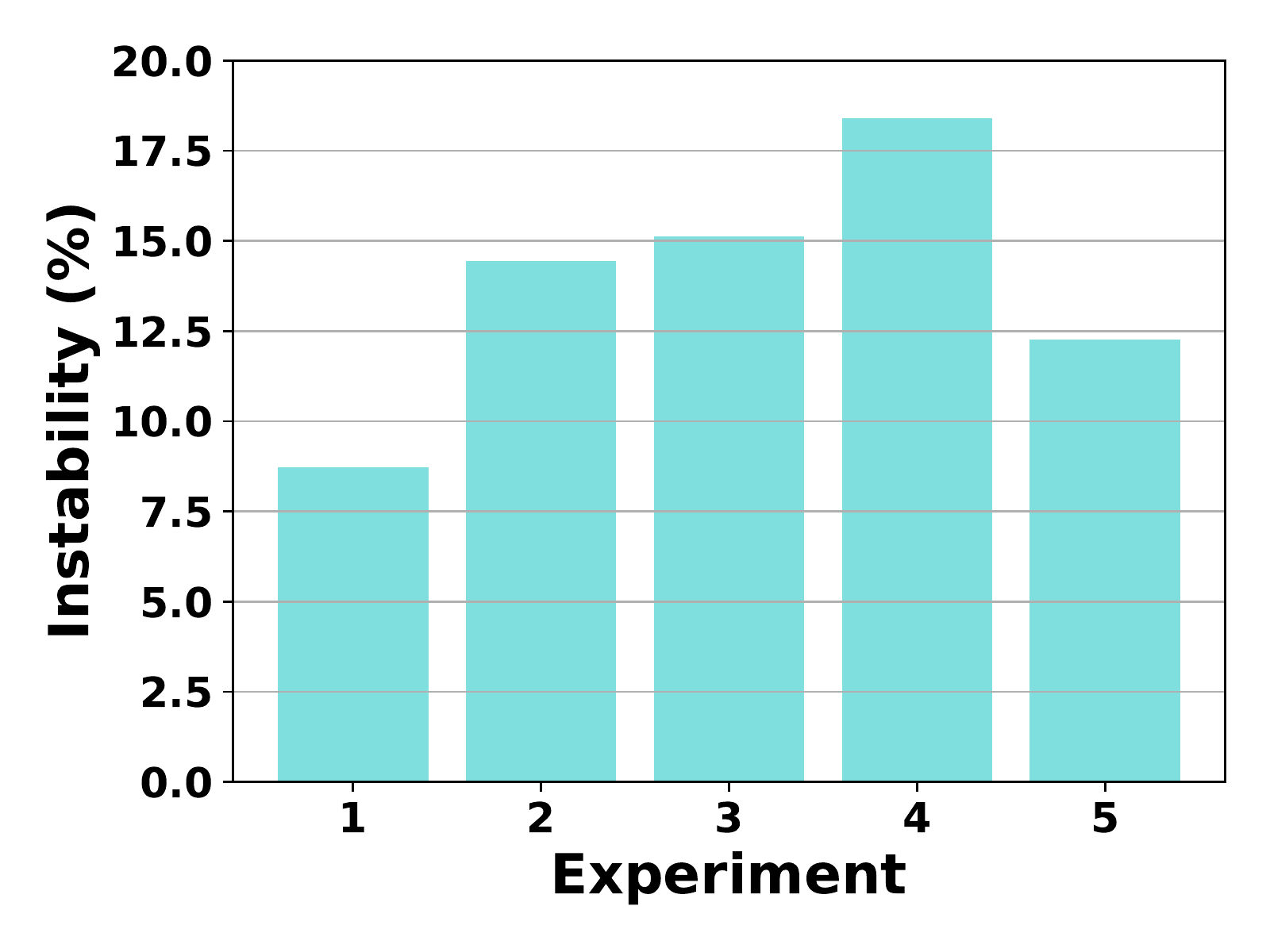}}\hfill
\centering
\subfigure[Instability over repeat photos of the same object with same phone.]{\label{fig:old-experiment-self-instability}
\includegraphics[width=0.47\columnwidth]{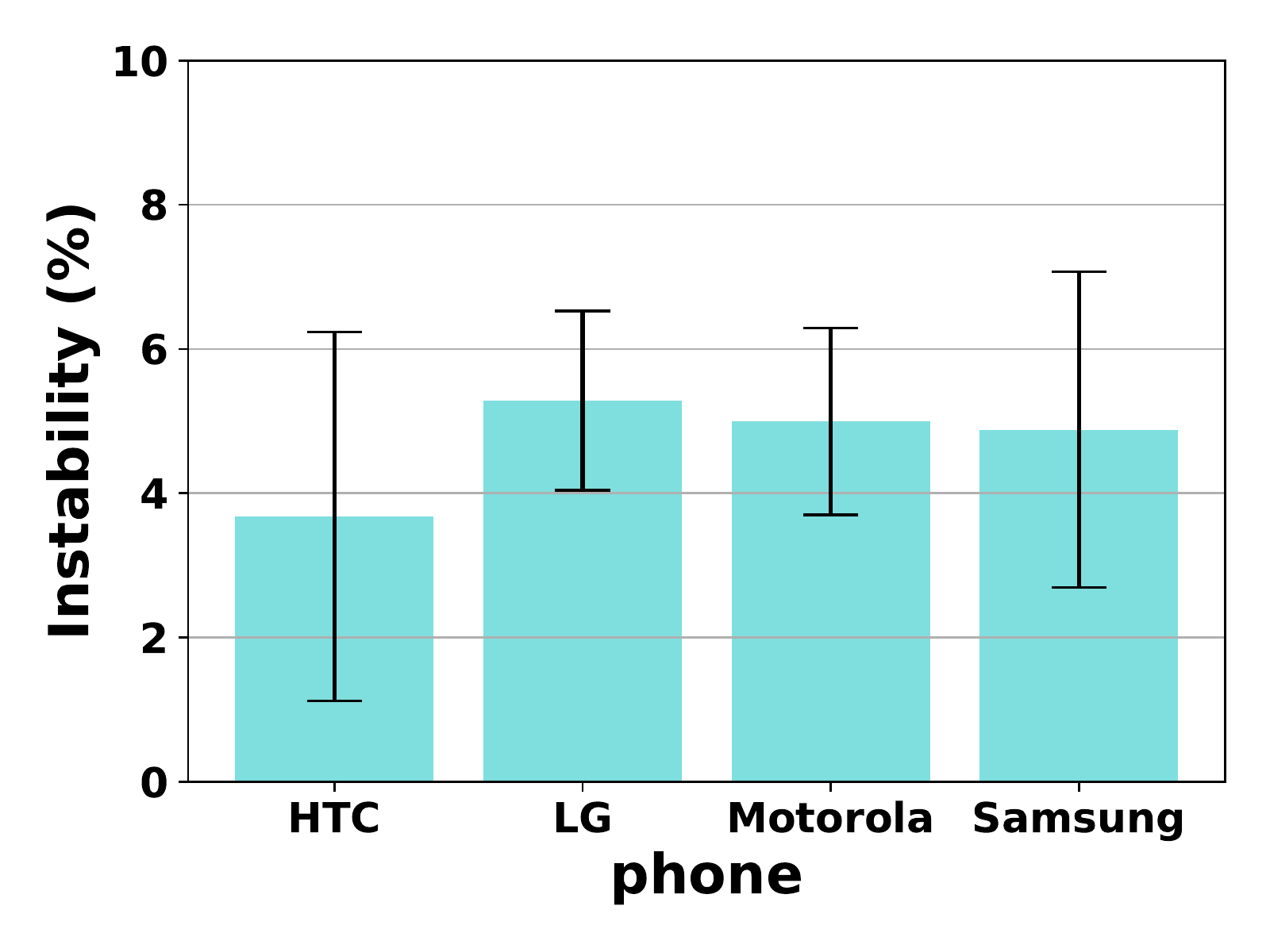}}
\vspace{-2.8pt}
\caption{End-to-end accuracy and instability. Accuracy does not capture the variability in predictions across different models.\label{fig:end-to-end-overview}}
\end{figure}
In our first set of experiments, we seek to quantify the amount of instability across classification tasks within the same phone model and across phone models, and to evaluate whether accuracy is a sufficient metric for capturing the variability in classification across edge devices.

\subsection{Accuracy vs. Instability\label{sec:acc-instability}}

We first evaluated the accuracy of our models on all 5 phones. The results, presented in Figure~\ref{fig:old-experiment}, show that accuracy is generally similar across all phone photos, and ranged between $59\%-64\%$. 

The instability across all models is depicted in Figure~\ref{fig:old-experiment-instability}. Instability is measured as the percentage of photos, where at least one of the phones was correct (\eg  classified the right class) and at least one of the phones was incorrect, when taking a photo of the same image on the computer screen. The results show that while the accuracy remains relatively stable across the different phones, there is a high degree of instability: for most of the classes about 15\% of the images yield at least one correct and one incorrect classification.

The results also demonstrate a large degree of variance in the instability between the different classes; some of them are more prone to instability.
Importantly,  \emph{this variation is not captured by the accuracy of each class}.

Figure~\ref{fig:old-experiment-variation} plots the variation in instability across the five angles, from left to right, and shows that instability does vary somewhat based on the angle of the image.
Figure~\ref{fig:old-experiment-self-instability} shows that there is instability even across experiments (\ie different angles) within the same phone model. However, this instability is much lower than the instability across different models.


Based on the results of the experiments, we can conclude accuracy is not a good metric to capture the variability of how well models perform across different devices.

\begin{figure}[h]
    \centering
    \subfigure[Stable images.]{
    \includegraphics[width=0.45\columnwidth]{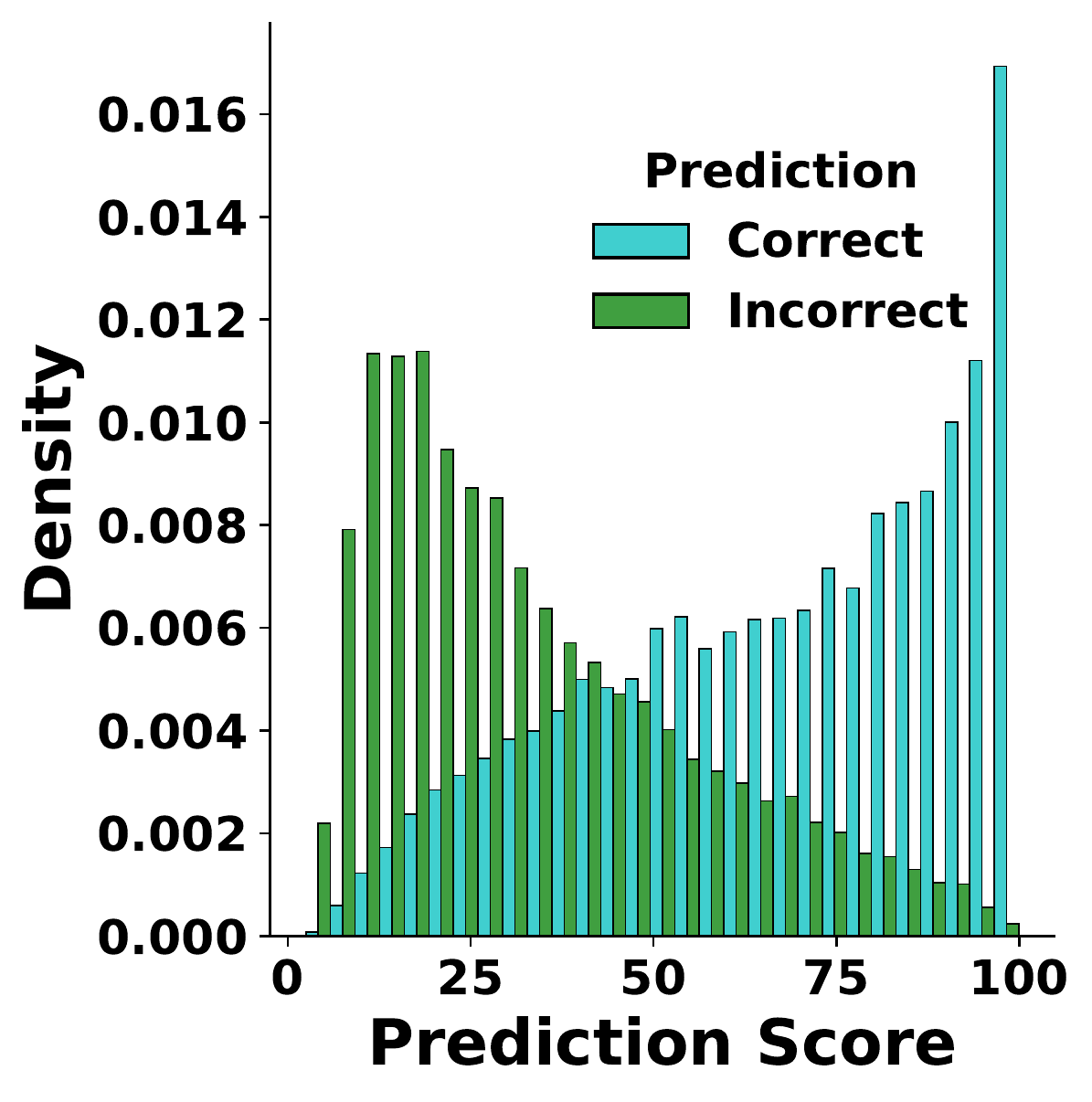} \label{fig:stable-images}}\hfill
    \subfigure[Unstable photos.]{
    \includegraphics[width=0.45\columnwidth]{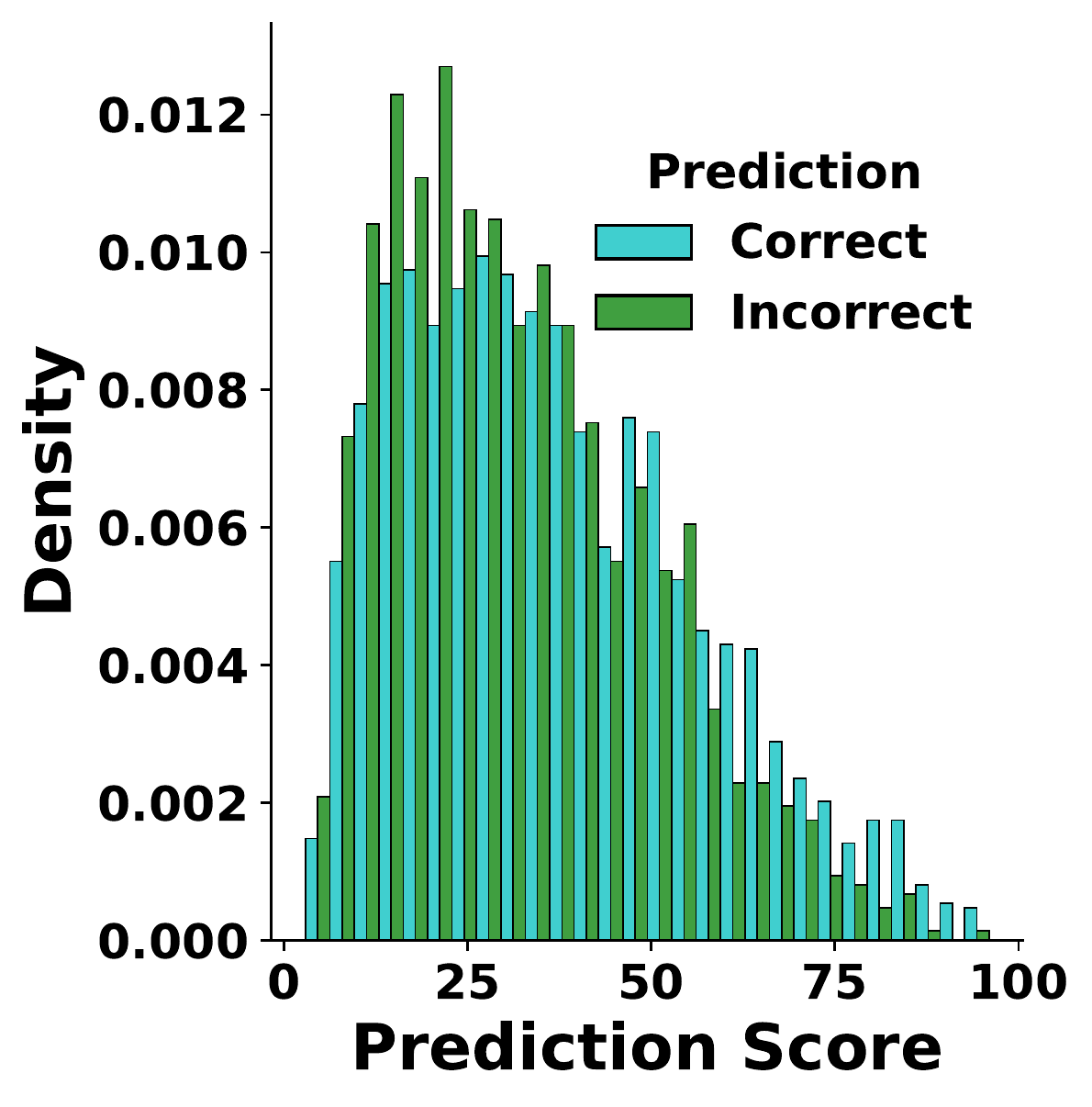} \label{fig:unstable-images}}
\vspace{-2.8pt}
\caption{Prediction score for stable and unstable images.}
\label{fig:app-experimet-confidence}
\end{figure}

\subsection{Model Confidence}
We evaluate the relationship between instability and the model's confidence in its prediction. Figure~\ref{fig:stable-images} shows the distribution of the prediction scores for stable photos (\ie those where all phones were correct or all were incorrect). We can see there is a clear correlation between the score and whether the model was correct or not.

Figure~\ref{fig:unstable-images} shows that for unstable predictions (\ie those where one phone was correct and one was incorrect), the prediction confidence of correct classifications tends to be almost identical to the confidence of the incorrect classifications. This implies that for most unstable images, the model has low confidence whereby even a small amount of noise can cause it to change its prediction.
Nevertheless, there is a noticeable group of outlier photos for which the model has very high confidence with one phone but low confidence with the other phone.
\section{Image Compression}
\label{sec:compression}

We now analyze the source and degree of instability within each phone, first focusing on compression. In this section, we analyze
only the raw photos taken in the end-to-end experiment on the iPhone and Samsung phone. The phones use two different compression schemes: Samsung uses JPEG and and iPhone uses HEIF. In order to isolate the effect of the ISP, we always use ImageMagick to compress and convert the photo to different formats.

\subsection{Compression Quality}
We compress the photos to JPEG with 3 different qualities: 100, 85 and 50. We used the compression parameter suggested by Google for machine learning~\cite{google-optimize-images}. 

\begin{table}[t!]
\caption{Accuracy and image size for different JPEG compression qualities.}
\begin{center}
\begin{small}
\begin{sc}
\begin{tabular}{llll}
\toprule
Metric & JPEG 100 & JPEG 85  & JPEG 50 \\
\midrule
Avg. Size [MB] & 3.05 & 0.65 & 0.25 \\
Accuracy & 54.0\% &  54.3\% & 54.5\% \\
\midrule
Instability & \multicolumn{3}{c}{7.6\%} \\
\bottomrule
\end{tabular}
\end{sc}
\end{small}
\end{center}
\label{tab:compression-quality}
\end{table}

Table~\ref{tab:compression-quality} shows the accuracy using different compression qualities. The results suggest that using Google's recommended compression parameters leads to very small differences in accuracy across compression qualities. In fact, in our experiment a higher compression ratio surprisingly yielded better accuracy. Yet the instability between the different qualities is 7.6\%.
Once again, despite the small difference in accuracy across the compression qualities, there is a noticeable instability. 

\begin{figure*}[t]
\begin{multicols}{4}%
\noindent%
{%
\subfigure[HEIF: backpack]{\includegraphics[width=0.23\textwidth, height=1.7cm]{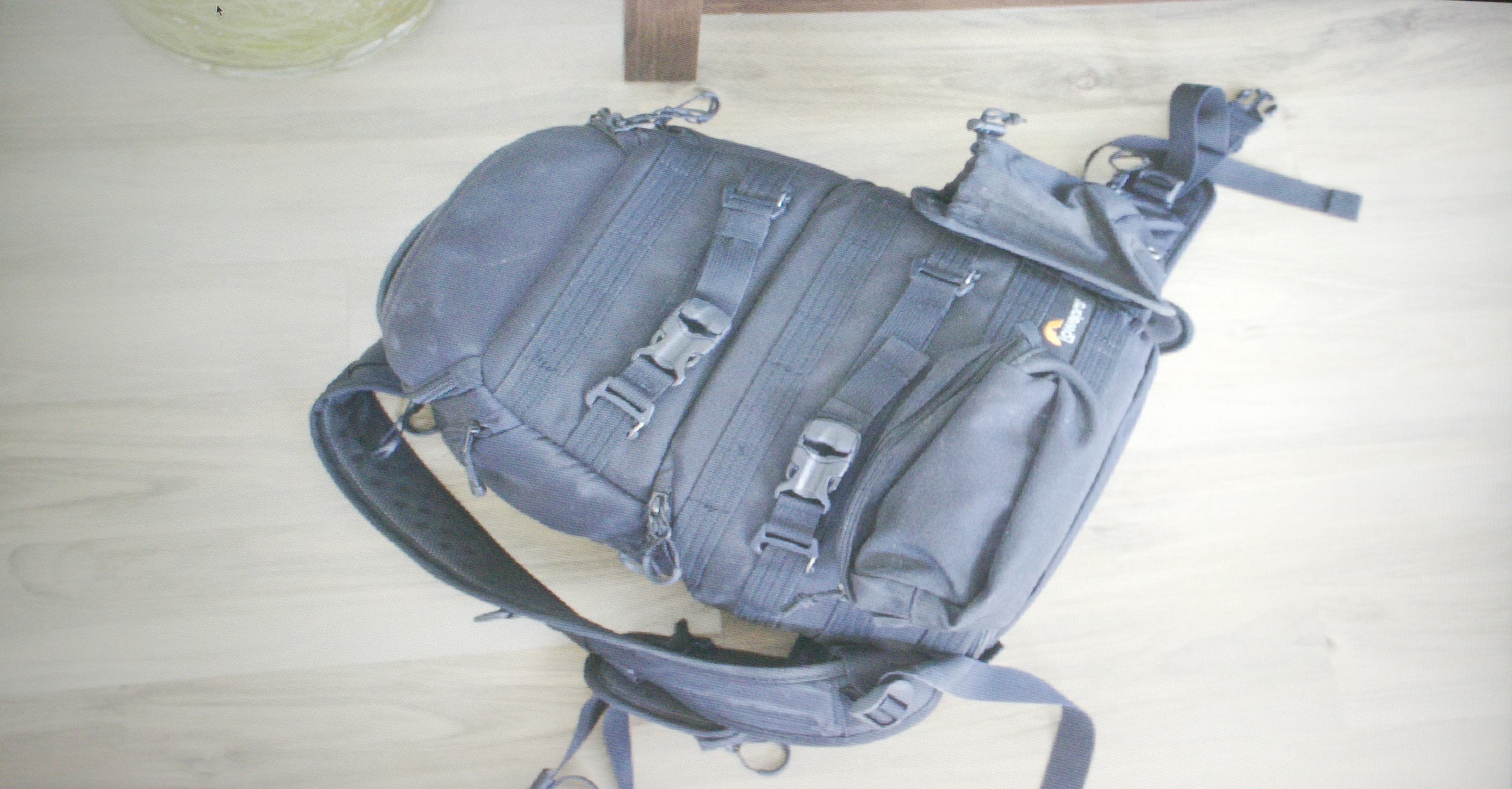}}\\%
\subfigure[JPEG: bonnet]{\includegraphics[width=0.23\textwidth, height=1.7cm]{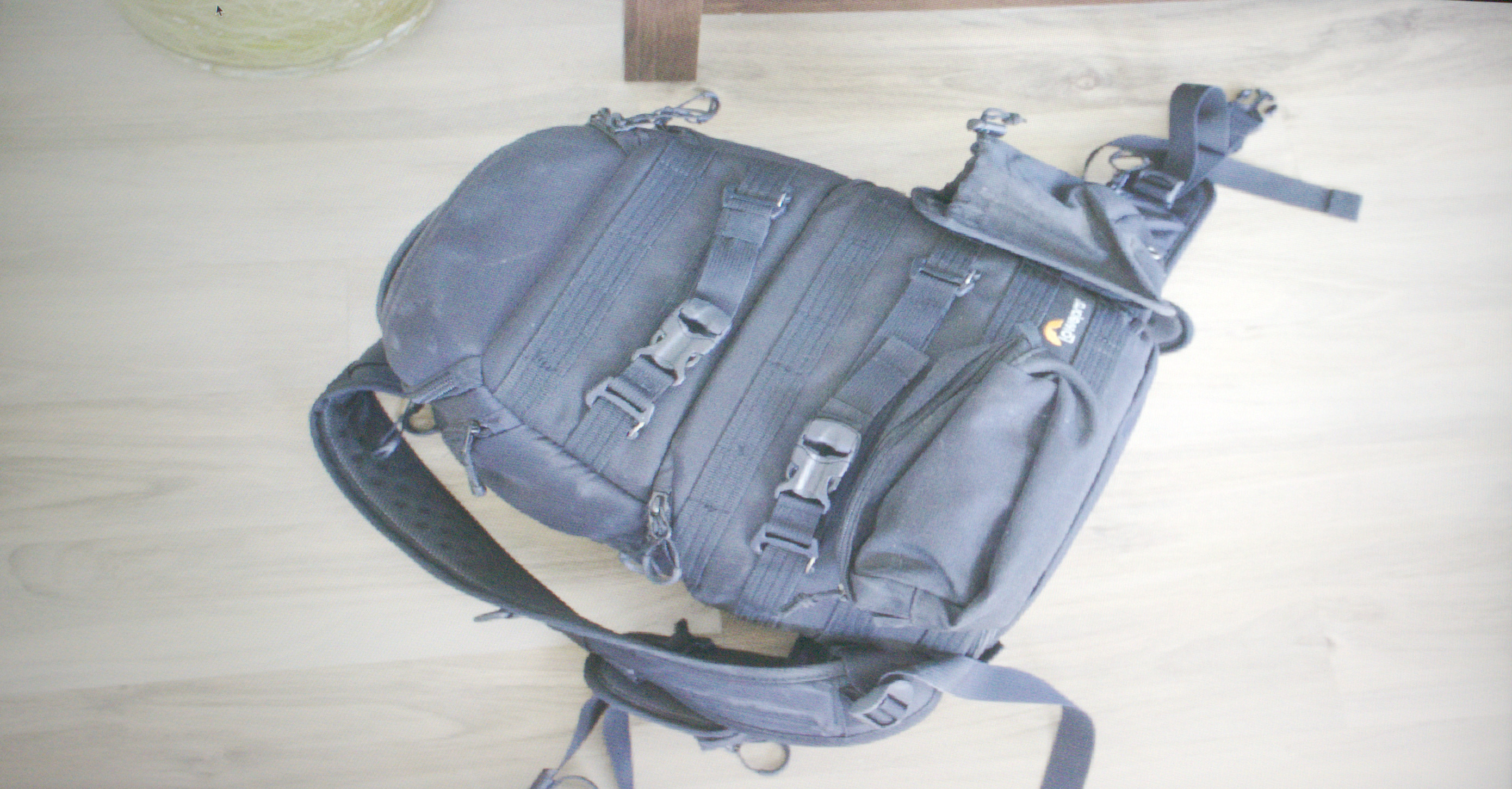}}\\%
\subfigure[PNG: bonnet]{\includegraphics[width=0.23\textwidth, height=1.7cm]{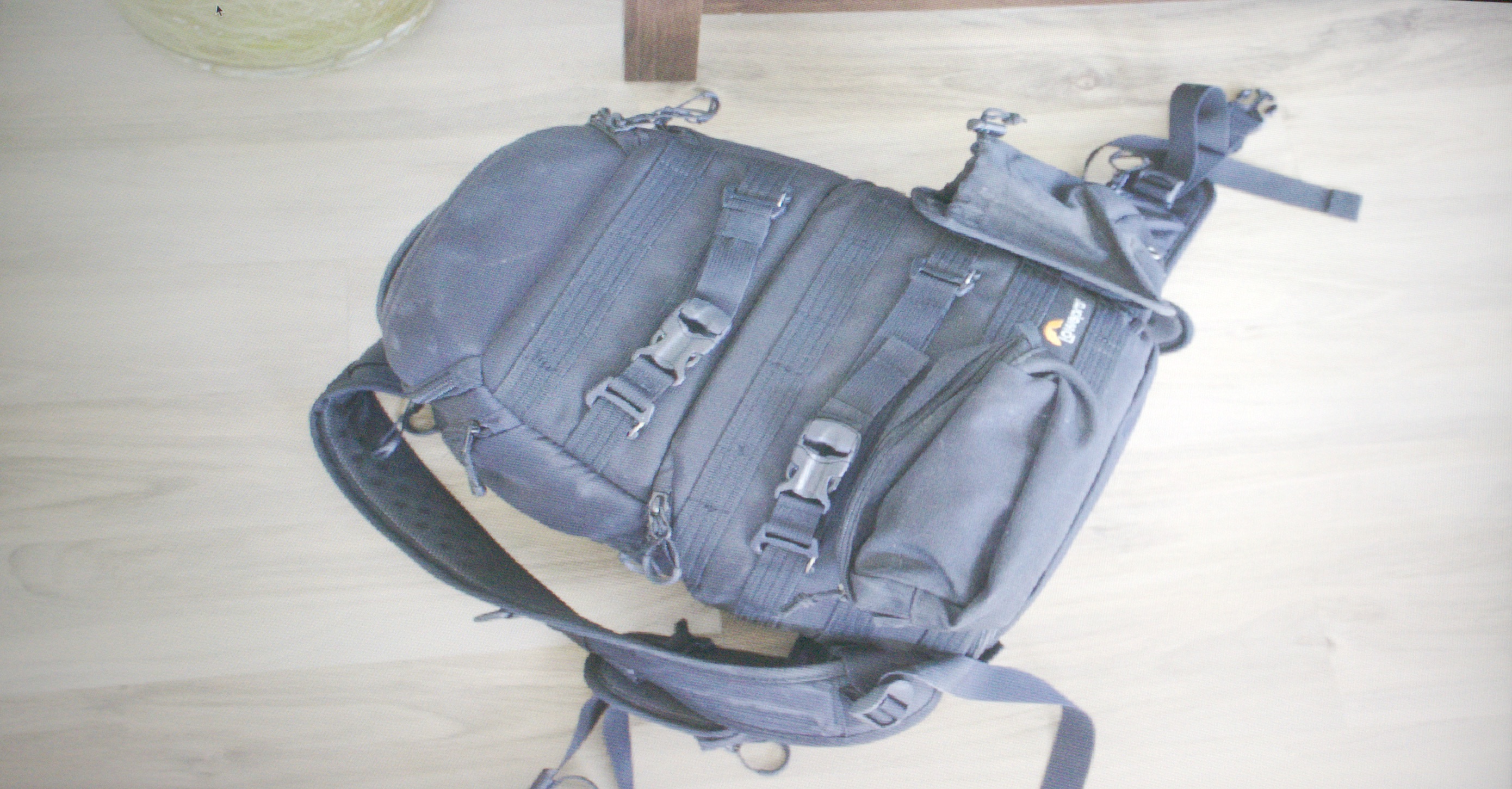}}\\%
}%
{%
\subfigure[HEIF: wine bottle]{\includegraphics[width=0.23\textwidth, height=1.7cm]{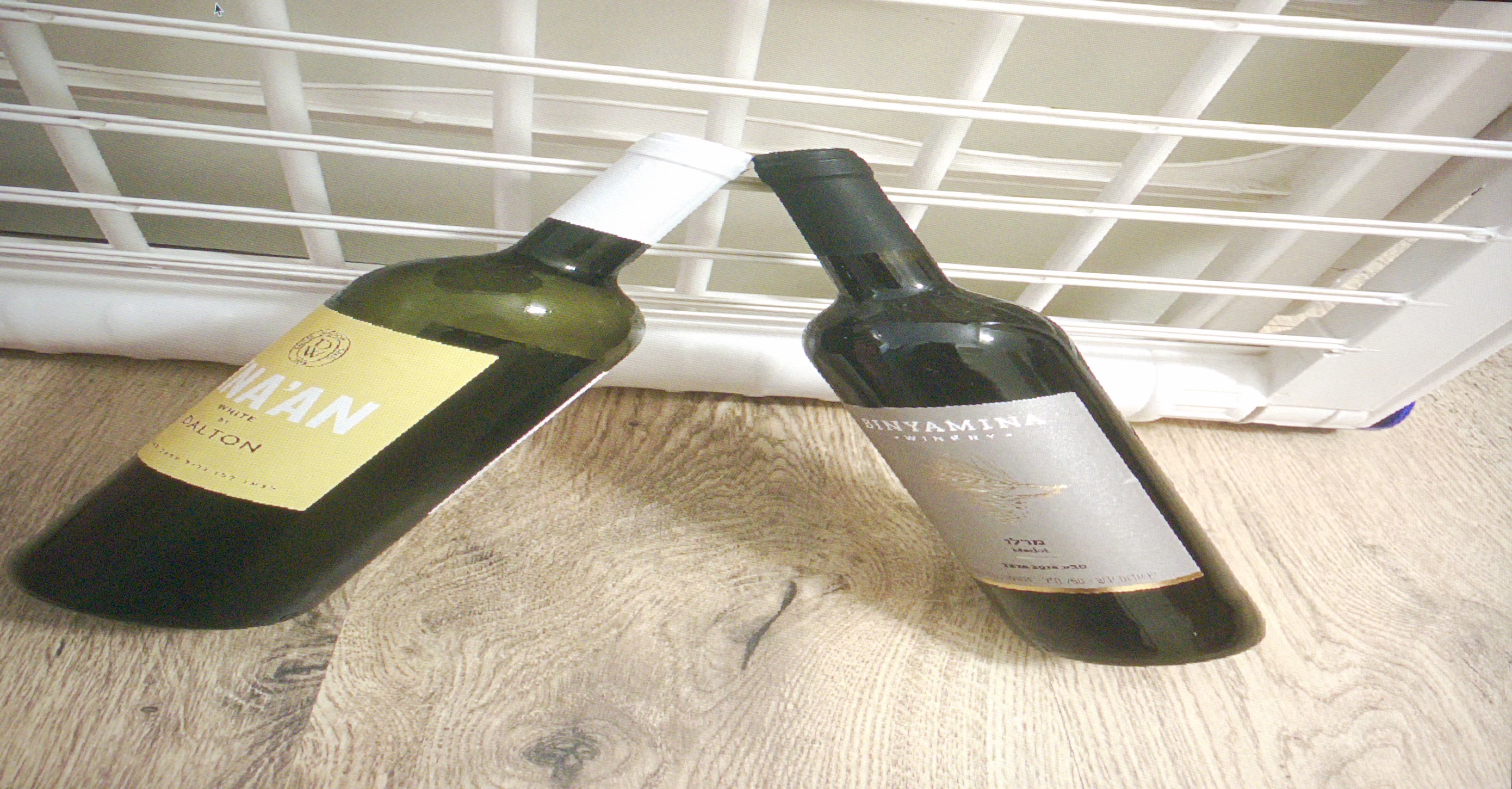}}\\%
\subfigure[JPEG: wine bottle]{\includegraphics[width=0.23\textwidth, height=1.7cm]{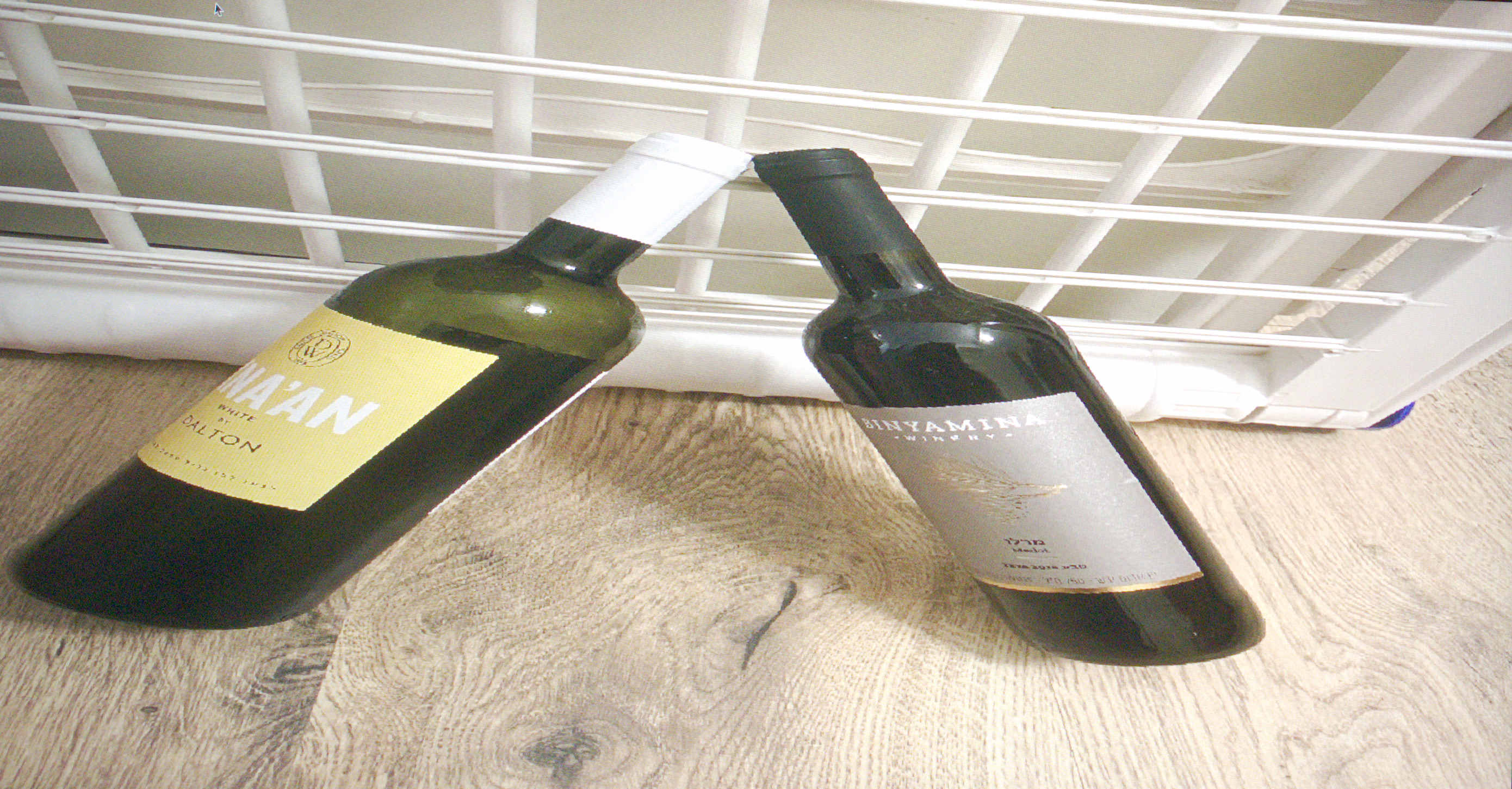}}\\%
\subfigure[PNG: beer bottle]{\includegraphics[width=0.23\textwidth, height=1.7cm]{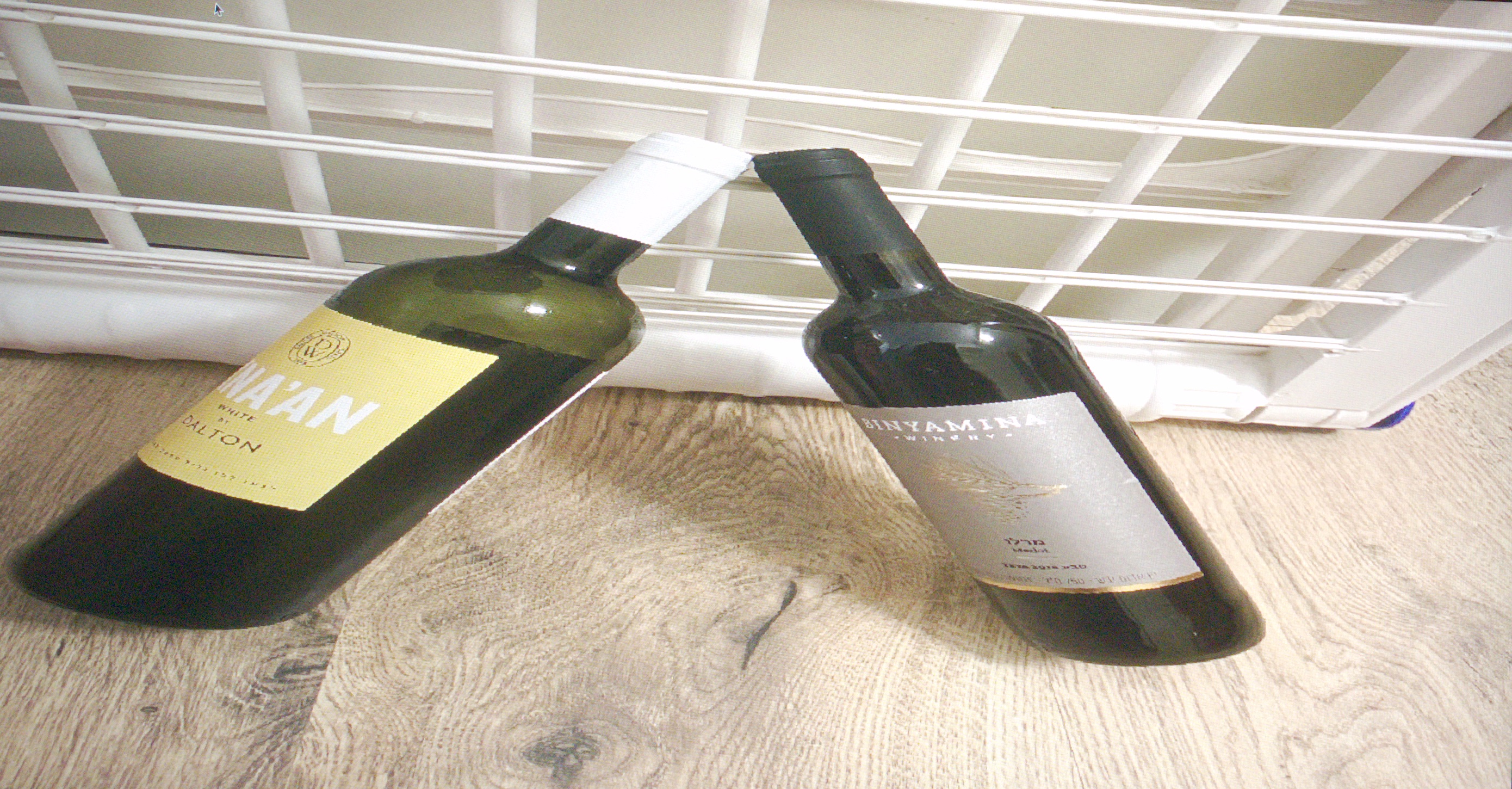}}\\%
}%
{%
\subfigure[HEIF: safety pin]{\includegraphics[width=0.23\textwidth, height=1.7cm]{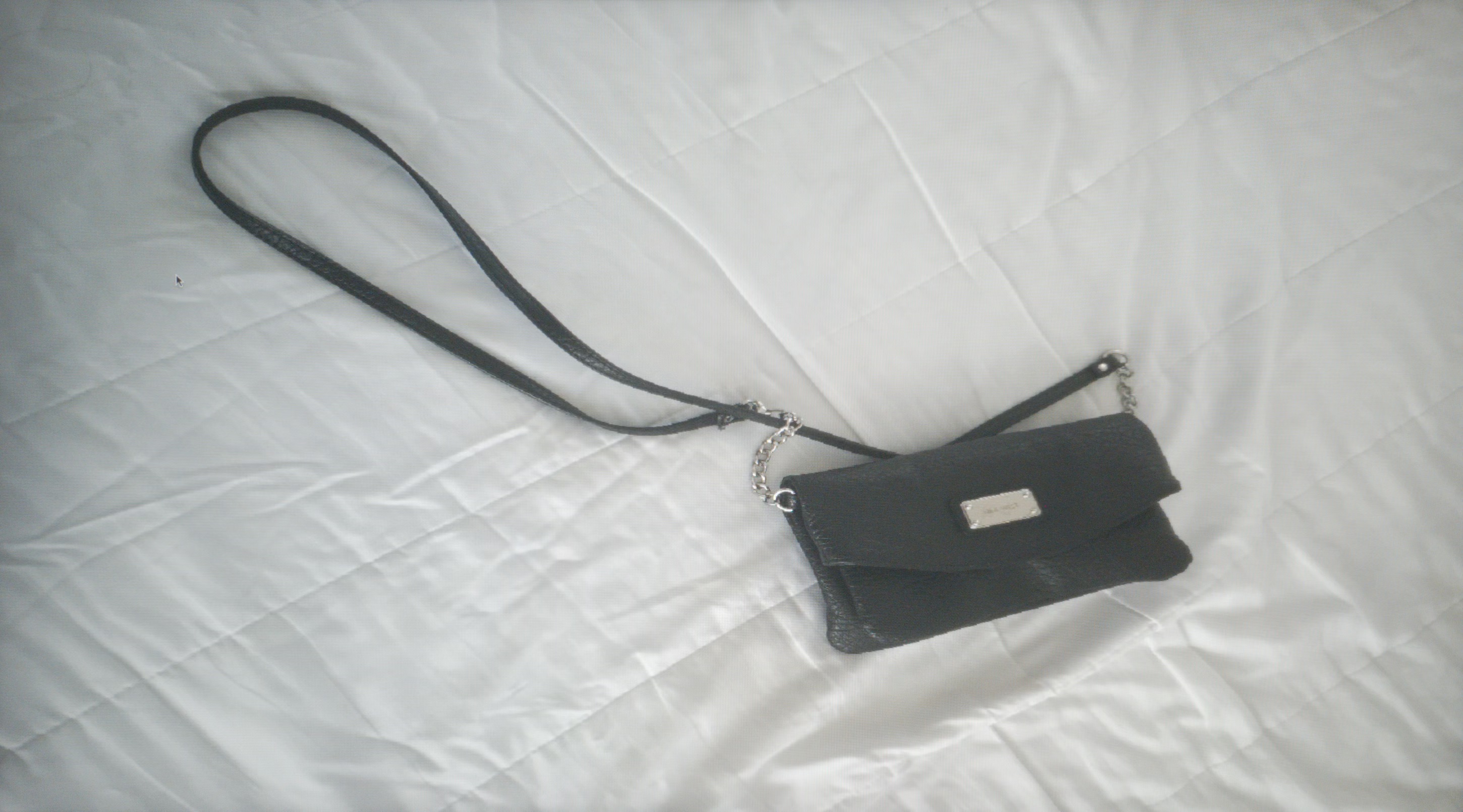}}\\%
\subfigure[JPEG: purse]{\includegraphics[width=0.23\textwidth, height=1.7cm]{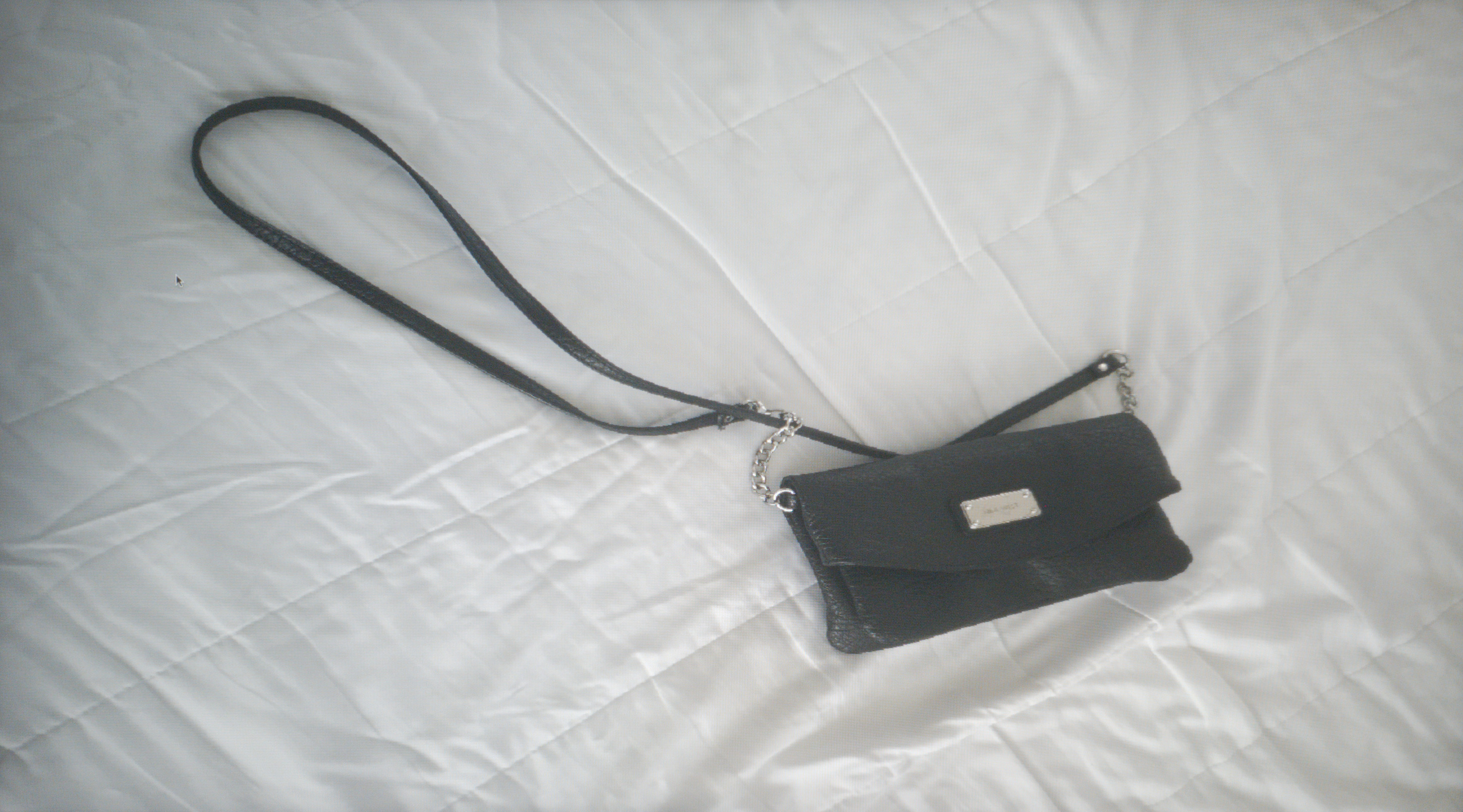}}\\%
\subfigure[PNG: stethoscope]{\includegraphics[width=0.23\textwidth, height=1.7cm]{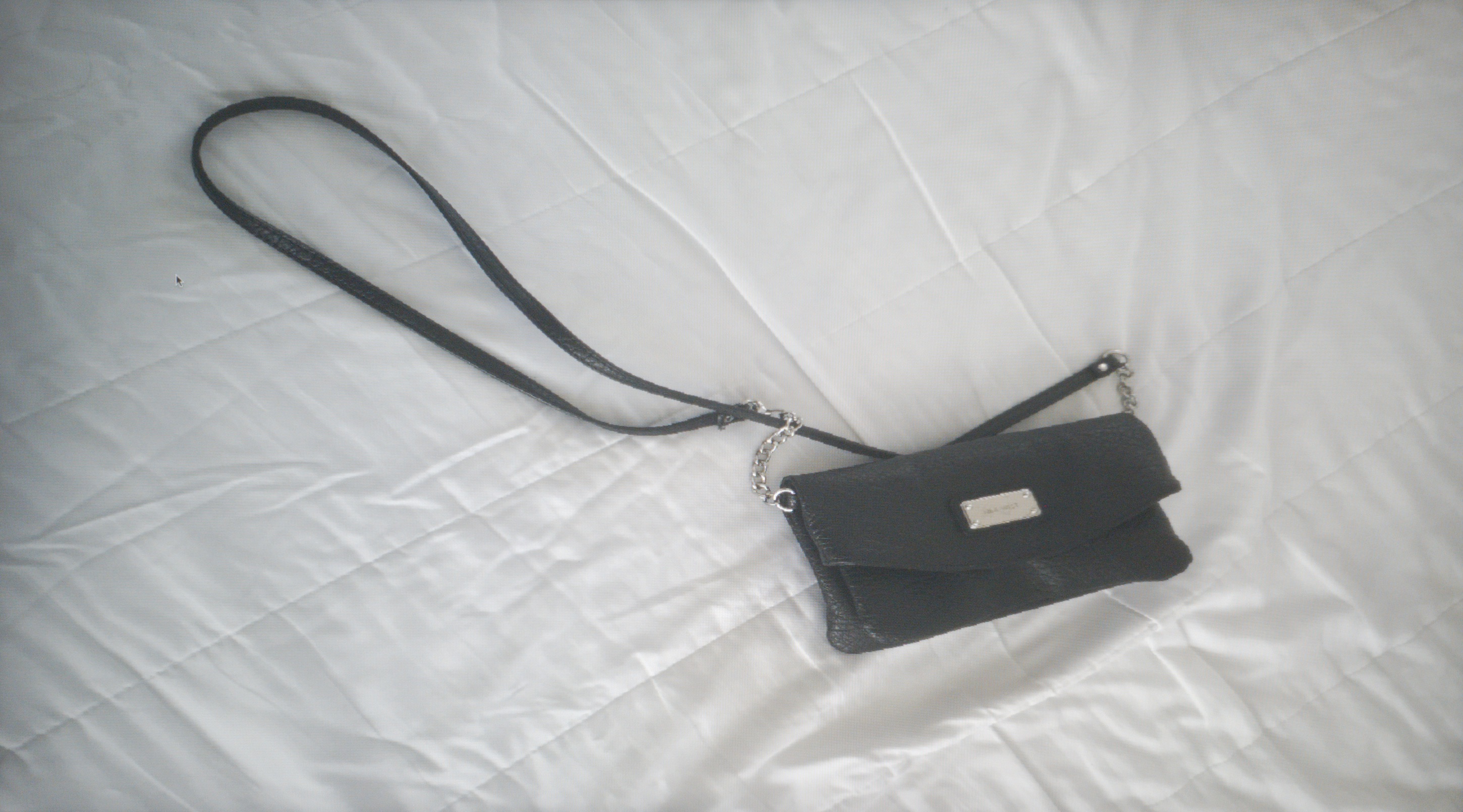}}\\%
}%
{%
\subfigure[JPEG-q100: beer bottle]{\includegraphics[width=0.23\textwidth, height=1.7cm]{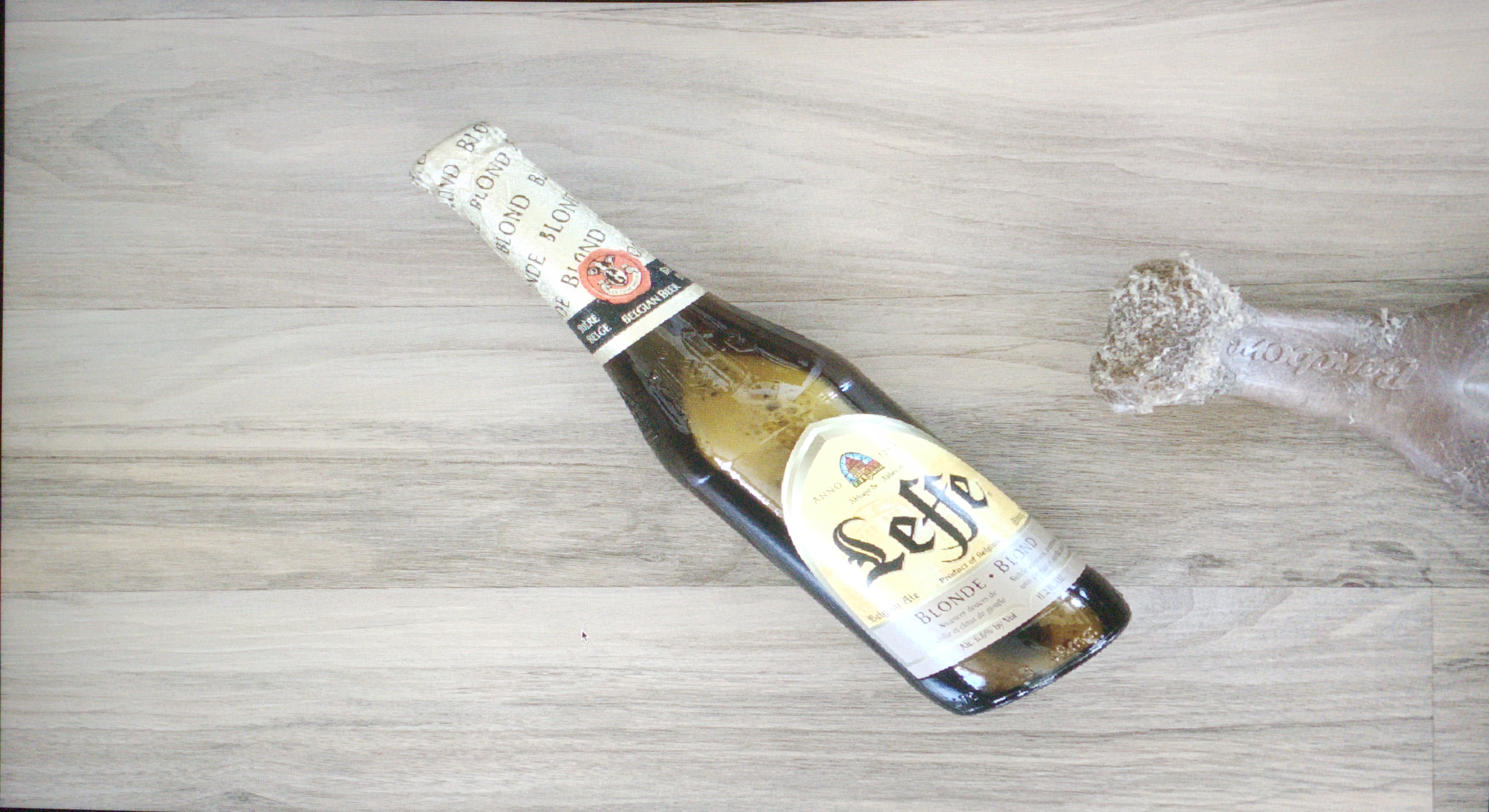}}\\%
\subfigure[JPEG-q85: lighter]{\includegraphics[width=0.23\textwidth,height=1.7cm]{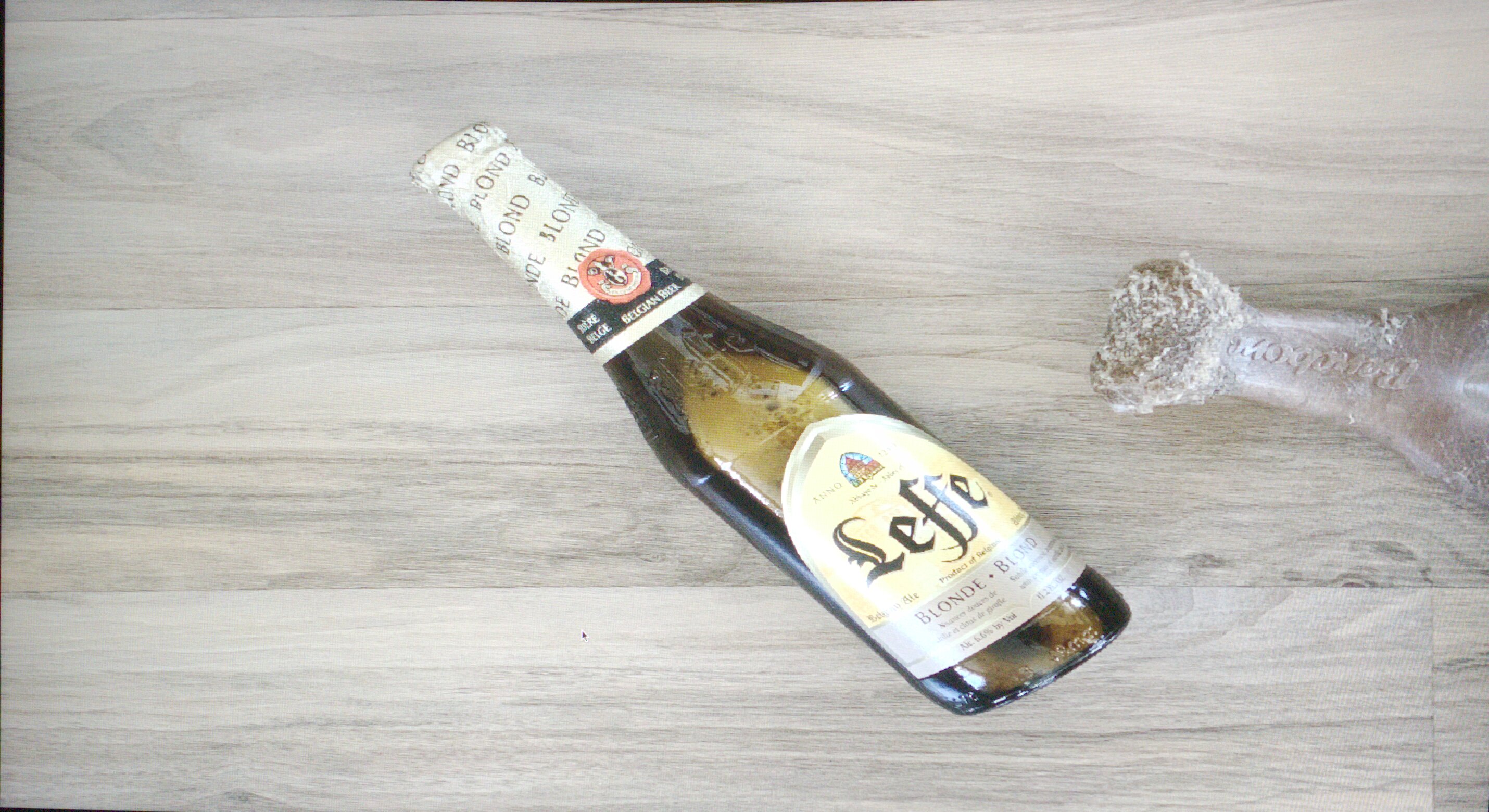}}\\%
\subfigure[JPEG-q50: lighter]{\includegraphics[width=0.23\textwidth, height=1.7cm]{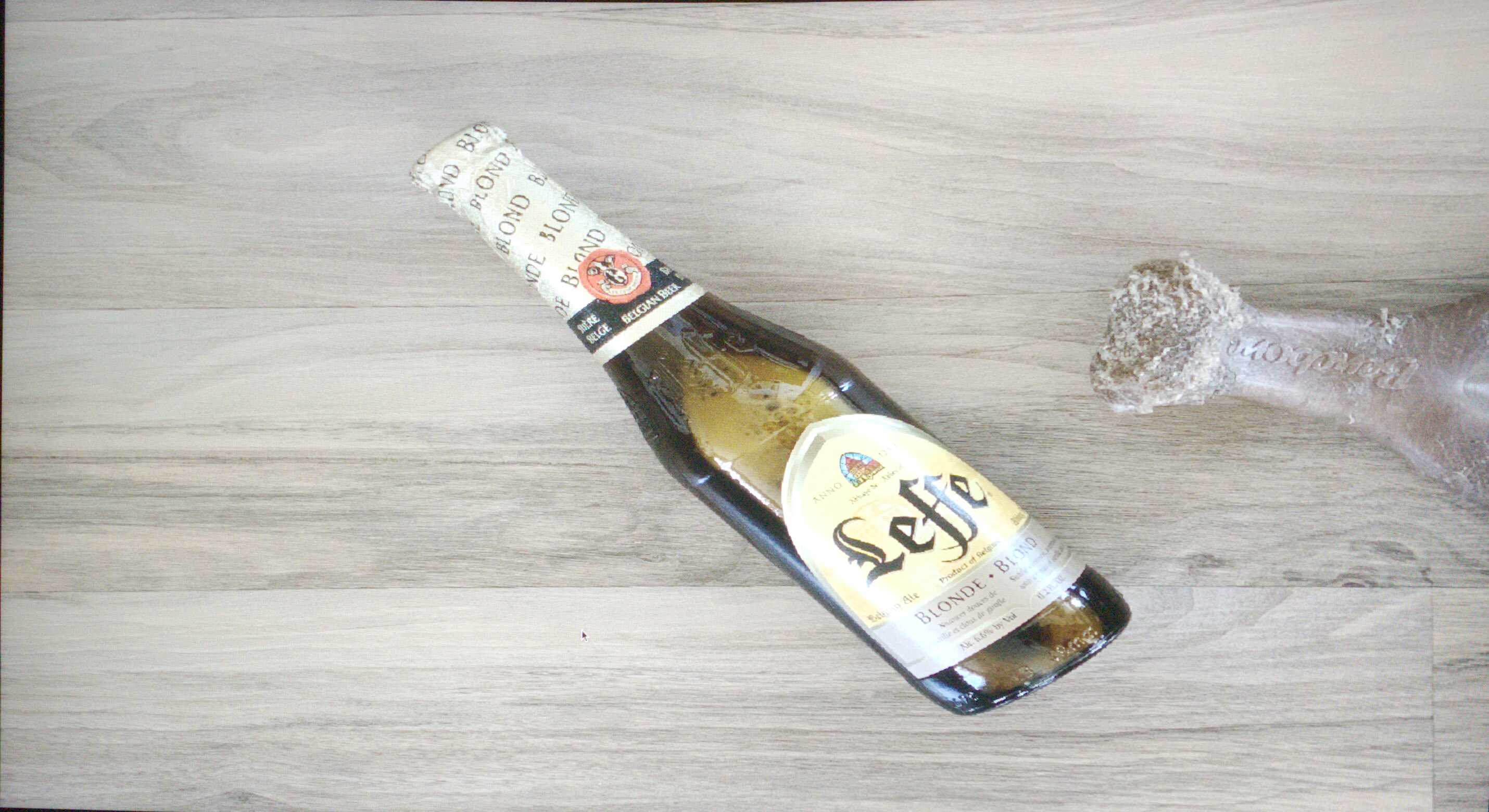}}%
}%
\end{multicols}
\vspace{-2.8pt}
\caption{Example of photos that cause instability due to different compression formats.\label{fig:compression-format-example}}
\end{figure*}

\subsection{Different Compression Formats}

We repeat this experiment but this time we compress the raw images into different formats: JPEG, PNG, WebP and HEIF.  Each format uses its default compression parameters.

\begin{table}[h]
\caption{Image size, accuracy and instability for different compression formats.}
\label{tab:compression-format}
\vskip 0.15in
\begin{center}
\begin{small}
\begin{sc}
\begin{tabular}{lllll}
\toprule
Metric & JPEG & PNG  & WebP & HEIF \\
\midrule
Avg. Size [MB] & 1.54 &  6.49 & 0.29 & 0.57\\
Accuracy & 53.9\% & 53.9\% & 55.2\% & 54.4\% \\
\midrule
Instability & \multicolumn{4}{c}{9.66\%} \\
\bottomrule
\end{tabular}
\end{sc}
\end{small}
\end{center}
\vskip -0.1in
\end{table}

In Table~\ref{tab:compression-format} we can see that as before, the different compression format led to small differences in accuracy. Yet the instability across formats is 9.66\%. Figure~\ref{fig:compression-format-example} shows a few examples of instability across the different compression formats. Notably the images are almost identical to the naked eye, but produce different model predictions.
\section{Image Signal Processor}
\label{sec:ISP}

An image signal processor (ISP) is a specialized image processor in digital cameras. The goal of the ISP in a phone camera is to take the raw sensor data from the camera, and transform it to a user-visible image. Common stages of an ISP pipeline include color correction, lens correction, demosaicing and noise reduction~\cite{buckler2017reconfiguring}.

Many mobile device producers utilize third-party ISP chips for at least part of their ISP pipelines. This implies that different phones, even if they have similar cameras, might produce very different photos. In fact it has been reported that phones from the same vendor might have different ISPs for the same model~\cite{phones-different-countries,phones-different-vendor}. Moreover, ISP pipelines in mobile devices have become more complex and might make different decisions on a photo based on the environment and the object being photographed~\cite{apple-isp,google-isp}. The phone vendor typically does not provide access to raw sensor data before the ISP chips, and so even when phones allow to access to raw photos, those photos might have gone through some ISP pipeline stages.

Past work has shown that changes to the ISP can have a significant effect on deep learning efficacy~\cite{buckler2017reconfiguring, liu2015ultra}. Therefore, we would like to characterize how differences in phone ISPs contribute to instability. To the best of our knowledge, no phone provides direct access to its full ISP pipeline. For this reason we used a technique employed in prior work of using ImageMagick and Adobe Photoshop as simulated software ISPs~\cite{buckler2017reconfiguring}.

We converted the raw photos taken from the iPhone and Samsung phones, using a software pipeline. We then evaluated the classification on the resulting uncompressed PNG. The results are shown in Table~\ref{tab:isp}.

\begin{table}[h]
\caption{Accuracy and instability for images converted with ImageMagick or Adobe Photoshop}
\label{tab:isp}
\vskip 0.15in
\begin{center}
\begin{small}
\begin{sc}
\begin{tabular}{ll}
\toprule
metric & result \\
\midrule
Adobe Accuracy & 49.96\% \\
ImageMagick Accuracy & 54.75\% \\
Instability & 14.11\%\\
\bottomrule
\end{tabular}
\end{sc}
\end{small}
\end{center}
\vskip -0.1in
\end{table}

We can see that different ISPs resulted in 14\% instability, implying that a large part of instability we saw in the end-to-end experiment may be attributed to ISP differences. Later in (\S\ref{sec:using-raw}) we will examine the potential of using raw images to overcome ISP and compression differences.
\section{Processor and OS}
\label{sec:OS}

In this section, we investigate two other potential causes for instability: (1) OS differences between the phone might affect how images are loaded and operations are scheduled; and (2) hardware differences might affect floating point calculations and instruction scheduling at inference time.

In order to test these two potential sources of instability, instead of taking images with the different phones and observing the differences, we run an experiment with a pre-defined set of photos, where we simply try to conduct inference on different phone models.
To this end we wrote a simple app that loads a subset of the Caltech101 dataset~\cite{caltech101} and runs classification on a subset of the images using MobileNetV2 trained on ImageNet. We tested our app on 5 phones, shown in Table~\ref{tab:firebase-phone-details}, using the Firebase Test Lab~\cite{firebase}, a service that allows developers to test their apps on different phones.

\begin{table}[h]
\caption{Phones used in the Firebase Test experiment.}
\label{tab:firebase-phone-details}
\vskip 0.15in
\begin{center}
\begin{small}
\begin{sc}
\begin{tabular}{ll}
\toprule
Phone & SoC \\
\midrule
Samsung Galaxy Note8 & Exynos 9 Octa 8895 \\
Huawei Mate RS & HiSilicon KIRIN 970 \\
Pixel 2 & Snapdragon 835 \\
Sony XZ3 & Snapdragon 845 \\
Xiaomi MI 8 Pro & Helio G90T (MT6785T) \\
\bottomrule
\end{tabular}
\end{sc}
\end{small}
\end{center}
\vskip -0.1in
\end{table}

During our experiment we observed very little instability. In fact the only two phones to produce any different predictions or confidences were the Huawei and Xiaomi Phones. Both Xiaomi and Huawei produced the same exact confidences and prediction, and the rest of the phone models (Samsung, Pixel and Sony) produced consistent predictions as well, but different than the Xiaomi and Huawei phones. These difference resulted in a small amount of instability: $0.64\%$.

We suspect the reason for the difference in prediction was due to differences in the OS's handling of JPEG decoding, rather then differences in hardware. To confirm this we looked at the MD5 hash of the loaded JPEG images, and indeed Huawei and Xiaomi produced different MD5 hashes then the rest of the phone. We further confirmed this by running our experiment on PNG images; when running on PNG images we detected no instability across all phones.
Therefore, we conclude that the processor and OS are not a major source of instability in our experimental setup.


\section{Takeaways From the Experiments}
To summarize, the end-to-end experiment demonstrated there is significant instability in model prediction generated by mobile phones even when they are taking photos of the same object in the same environmental conditions. Our experimental results show that there can be 14\%-17\% instability overall and even higher for individual classes.

We examined potential root causes of the instability. From our analysis it seems that most of the instability can be explained by the different ways mobile phone process images. We have seen that compression differences can cause about 5-10\% instability, and ISP pipeline differences may cause about 14\%. We saw very little instability caused by running different OSes or processors.
In the next section, we evaluate how to mitigate instability across devices.

\section{Mechanisms to Reduce Instability}
In this section, we propose and evaluate three different approaches to reduce instability when running a model on different devices:
\begin{denseenum}
\item Fine-tuning the model using stability loss to be more robust to input noise.
\item Reducing noise in the input by using raw images, and applying consistent ISP and compression.
\item Modifying the prediction task to account for instability, \eg using the top 3 predicted classes.
\end{denseenum}

\subsection{Stability Training}

The main approach we investigated to reducing instability is fine-tuning the model to images taken by the phone. However, fine-tuning to a specific phone model introduces some challenges. A naive approach would be to train on photos taken from every phone the model might run on. However, there are thousands of constantly changing phone models, and developers often cannot anticipate which devices will run their model.

\begin{figure}[h]
    \centering
    \subfigure[Gaussian noise generator]{\label{fig:gaussian-stability}%
    \includegraphics[width=0.235\textwidth]{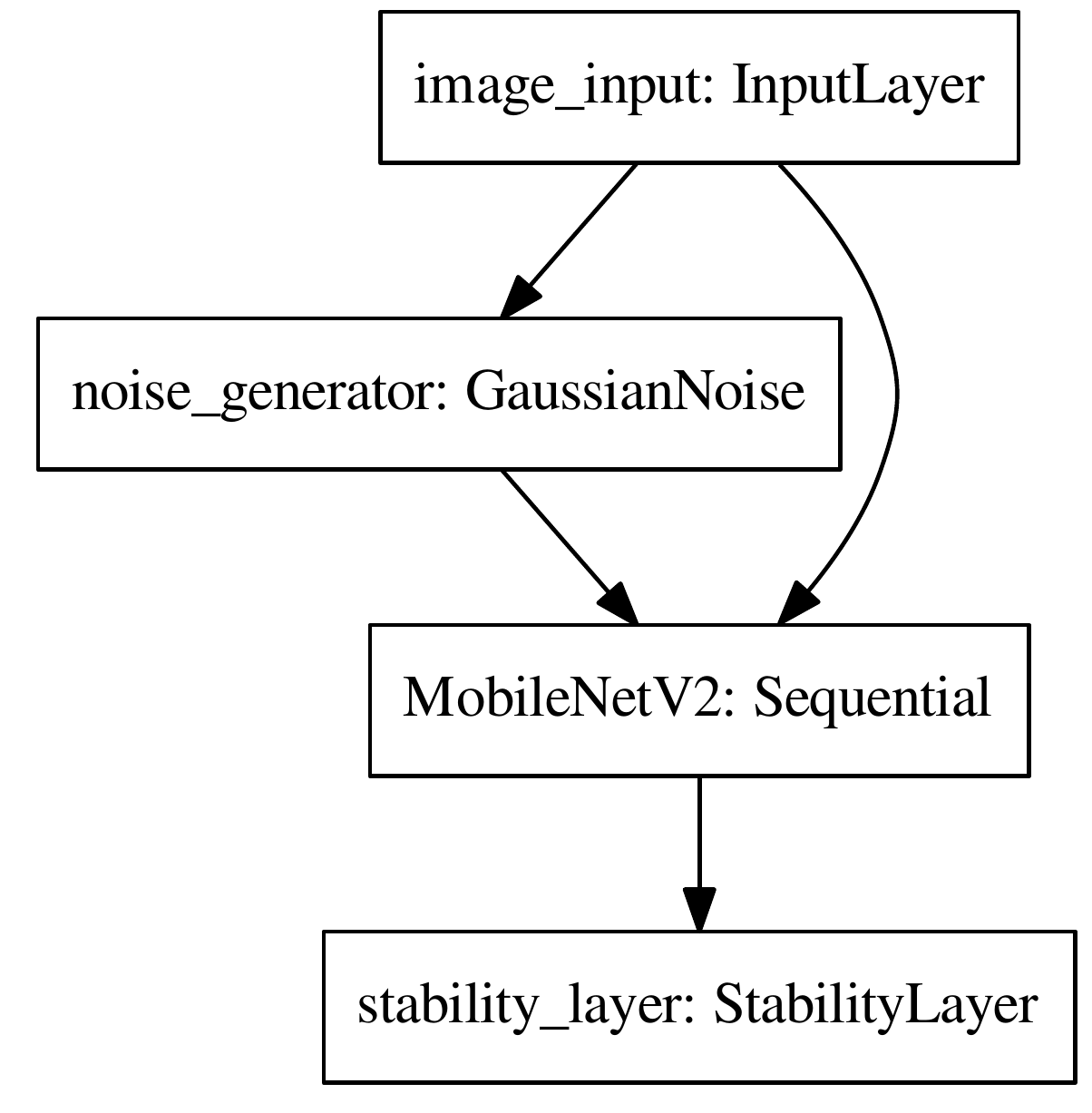}}\hfill
    \subfigure[Two images as inputs]{\label{fig:two-images-stability}%
    \raisebox{0.7cm}{%
    \includegraphics[width=0.235\textwidth]{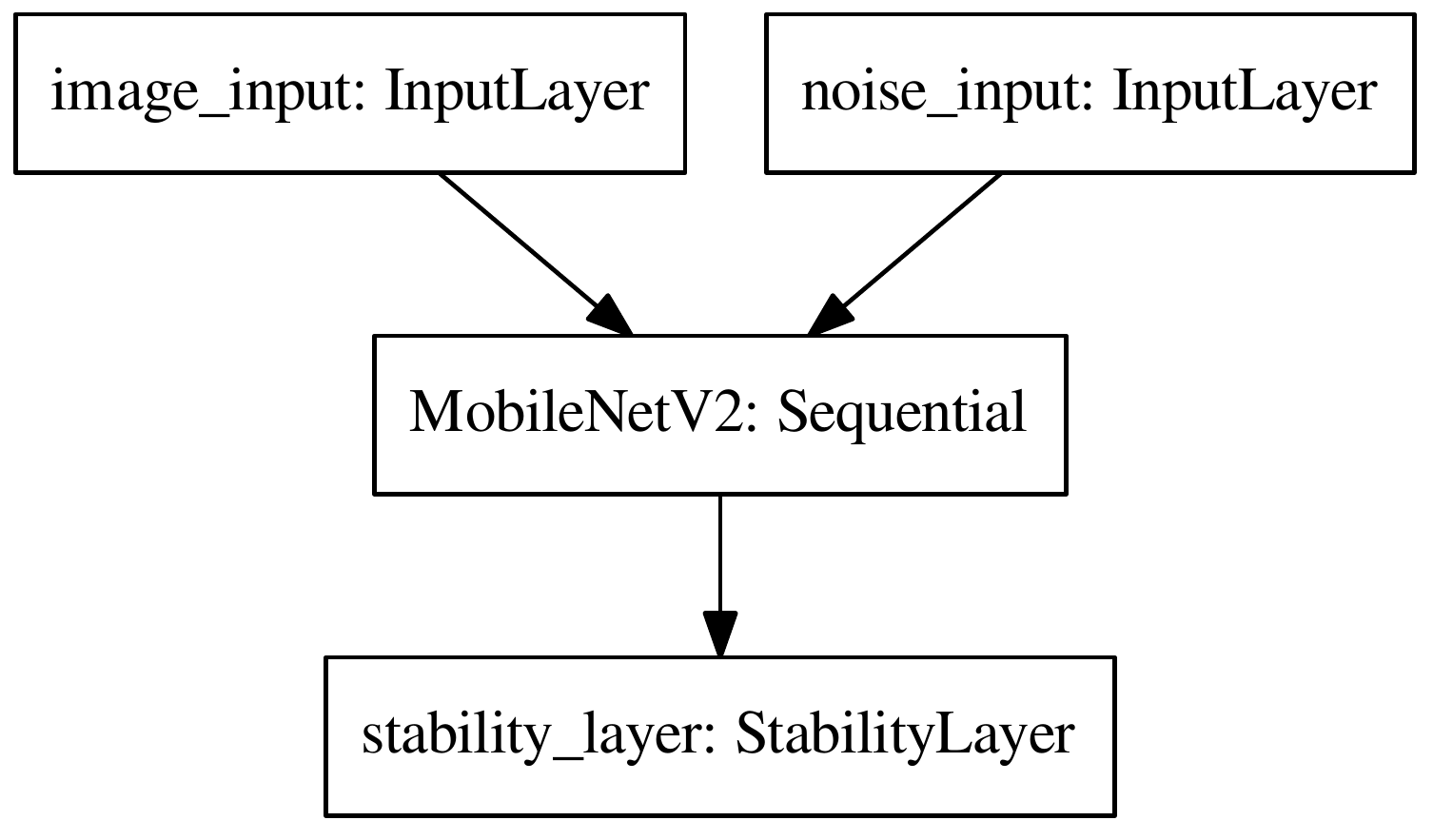}}}
    \vspace{-2.8pt}
    \caption{Examples of different versions of the fine-tuned stability model. We tested different versions of noise generation and stability loss. In the embedding distance stability loss an extra dense layer is added to the model.}
    \label{fig:stability-model}
\end{figure}

\begin{table*}[t!]
    \caption{Instability between iPhone and Samsung photos with fine-tuned MobileNetV2 with different stability losses and noise generation schemes. The hyper parameters for each training scheme are also included.}
    \label{tab:stability-training}
    \vskip 0.15in
    \begin{center}
    \begin{small}
    \begin{sc}
    \subfigure[Embedding distance loss.]{
    \begin{tabular}{ccc}
        \toprule
        Noise & Hyper Parameters & Instability \\
        \midrule
        Two Images & $\alpha=0.001$ & 3.91\% \\
        Subsample & $\alpha=0.001$ $\#images=10$ & 4.22\% \\
        Distortion & $\alpha=0.01$ & 5.12\% \\
        Gaussian & $\alpha=0.001$ $\sigma^{2}=0.04$ & 5.12\% \\
        No Noise & n/a & 7.22\% \\
        \bottomrule
    \end{tabular}\label{tab:embedding-instability}}
    \subfigure[Relative entropy loss.]{
    \begin{tabular}{ccc}
        \toprule
        Noise & Hyper Parameters & Instability \\
        \midrule
        Two Images & $\alpha=0.01$ & 6.32\% \\
        Subsample & $\alpha=0.01$ $\#images=10$ & 5.72\% \\
        Distortion & $\alpha=0.1$ & 4.52\% \\
        Gaussian & $\alpha=1$ $\sigma^{2}=0.025$ & 4.82\% \\
        No Noise & n/a & 6.62\% \\
        \bottomrule
    \end{tabular}\label{tab:kl-instability}}
    \end{sc}
    \end{small}
    \end{center}
    \label{tab:stability-results}
\end{table*}

\begin{figure*}
    \centering
    \subfigure[Embedding distance loss]{%
    \label{fig:embedding-pr}%
    \includegraphics[width=0.45\textwidth]{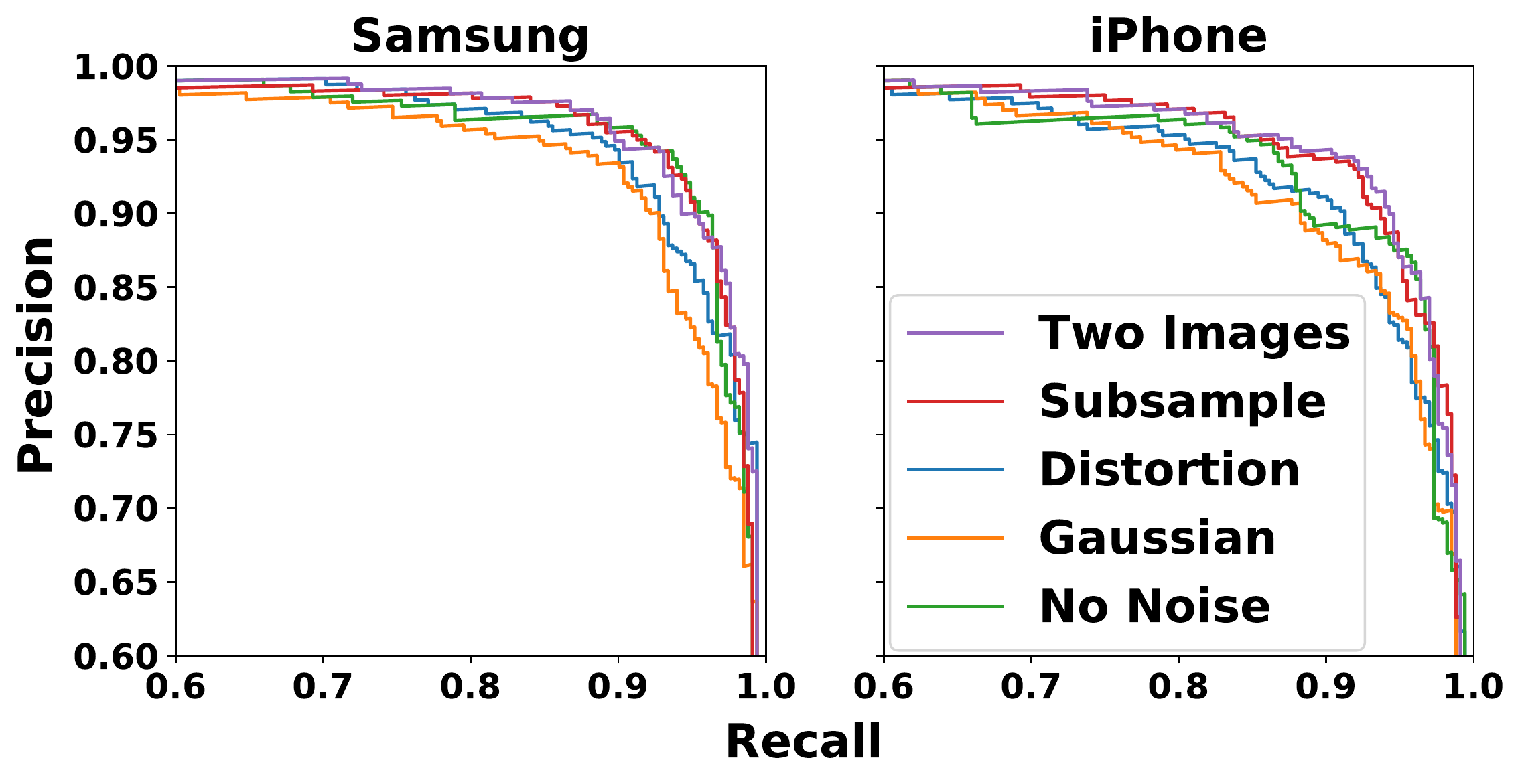}}
    \subfigure[Relative entropy loss]{%
    \label{fig:kl-stab-pr}%
    \includegraphics[width=0.45\textwidth]{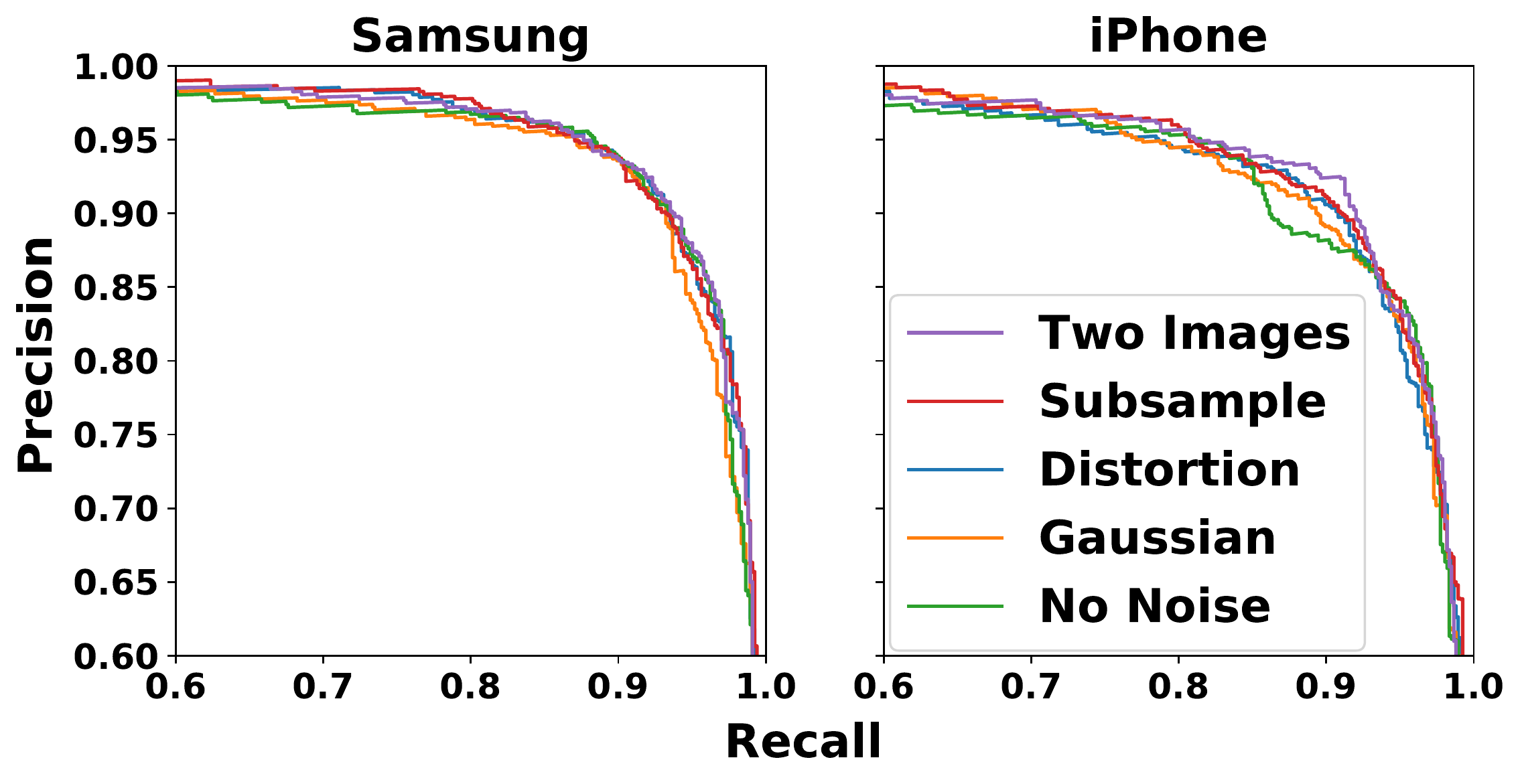}}
    \vspace{-0.25cm}
    \caption{Precision-recall curves for the different stability fine-tuning schemes, tested on Samsung and iPhone phones. Stability training not only reduces instability but slightly increases accuracy.}
    \label{fig:pr-curves}
\end{figure*}

Therefore, an alternative solution is to use techniques that make the model more robust to noise.
Robust machine learning is a well-researched area~\cite{NIPS2017_6821,nguyen2015deep,szegedy2013intriguing,kurakin2016adversarial,bastani2016measuring,cisse2017parseval,tanay2016boundary,survay}, but most papers deal with robustness to adversarial examples hand-crafted to change model predictions, rather then non-adversarial noise. We are inspired by prior work by Zheng \etal~\cite{StabilityTraining} on stability training, which is a technique on how to improve model accuracy under noise from the environment. We further develop the stability training idea to reduce instability caused by different devices. 

In stability training, during training each training image is complemented by an augmented image, which is generated by adding uncorrelated Gaussian pixel noise, \ie if $k$ is the pixel index for image $x$, and $\sigma^{2}$ is the standard deviation of the noise, we generate $x'$ such that:
\setlength{\belowdisplayskip}{1mm} \setlength{\belowdisplayshortskip}{1mm}
\setlength{\abovedisplayskip}{1mm} \setlength{\abovedisplayshortskip}{1mm}
$$x'_{k} = x_{k} + \epsilon,  \epsilon \sim \mathcal{N}(0,\sigma^2)$$
Because Gaussian noise is not representative of the differences between two phone images, we tried another version of noise generation: simulating the noise introduced by different phone ISPs. Our simulated phone noise randomly distorts different aspects of the training image: the hue, contrast, brightness, saturation and JPEG compression quality.

The model is trained with an augmented loss function, where $\theta$ are the trainable model weights, and $\alpha$ is an adjustable hyper parameter:
$$L(x,x',\theta) = L_{0}(x,\theta) + \alpha L_{s}(x,x',\theta)$$
$L_{0}$ is the regular classification cross entropy loss, \ie for the predicted label vector $y$ and true label vector $\hat{y}$:
$$L_{0}(x,\theta) = \unaryminus\sum\hat{y}_{j}Log(P(y_{j}|x,\theta))$$
The stability loss $L_{s}(x,x',\theta)$ can come in one of two forms:
\begin{denseitemize}
    \item The relative entropy (Kullback–Leibler divergence) over the model prediction of the input image against those of the noisy image:
$$L_{s}(x,x',\theta) = \unaryminus\sum P(y_{j}|x,\theta)Log\Big(\frac{P(y_{j}|x',\theta)}{P(y_{j}|x,\theta)}\Big)$$
    \item The Euclidean distance between the embedding layer (the input to the last fully-connected layer of the model) between the input image and the noisy image:
$$L_{s}(x,x',\theta) = \|f(x,\theta) - f(x',\theta)\|_{2}$$
\end{denseitemize}

We implemented the stability training technique~\cite{StabilityTraining} using Keras~\cite{keras} with a Tensorflow 2.3.0 backend~\cite{abadi2016tensorflow}. We tried different variations of the stability training model to evaluate which approach would reduce instability the most. An illustration of some of the different versions of stability models we tried can be seen in Figure~\ref{fig:stability-model}. We train using a Tesla K80 GPU.

We compare our noise generation models to a version of the stability training in which the noisy image is not automatically generated but rather supplied as a separate input to the model. For instance, if we are training on images from the end-to-end experiment taken by the Samsung phone, we can supply the equivalent images from the iPhone as the noisy version of the images.

\begin{figure*}[ht!]
\centering
\subfigure[JPEG vs. raw converted images on Samsung and iPhone.]{\label{fig:app-experiment-inconsistancy-a}%
\includegraphics[width=0.3\textwidth]{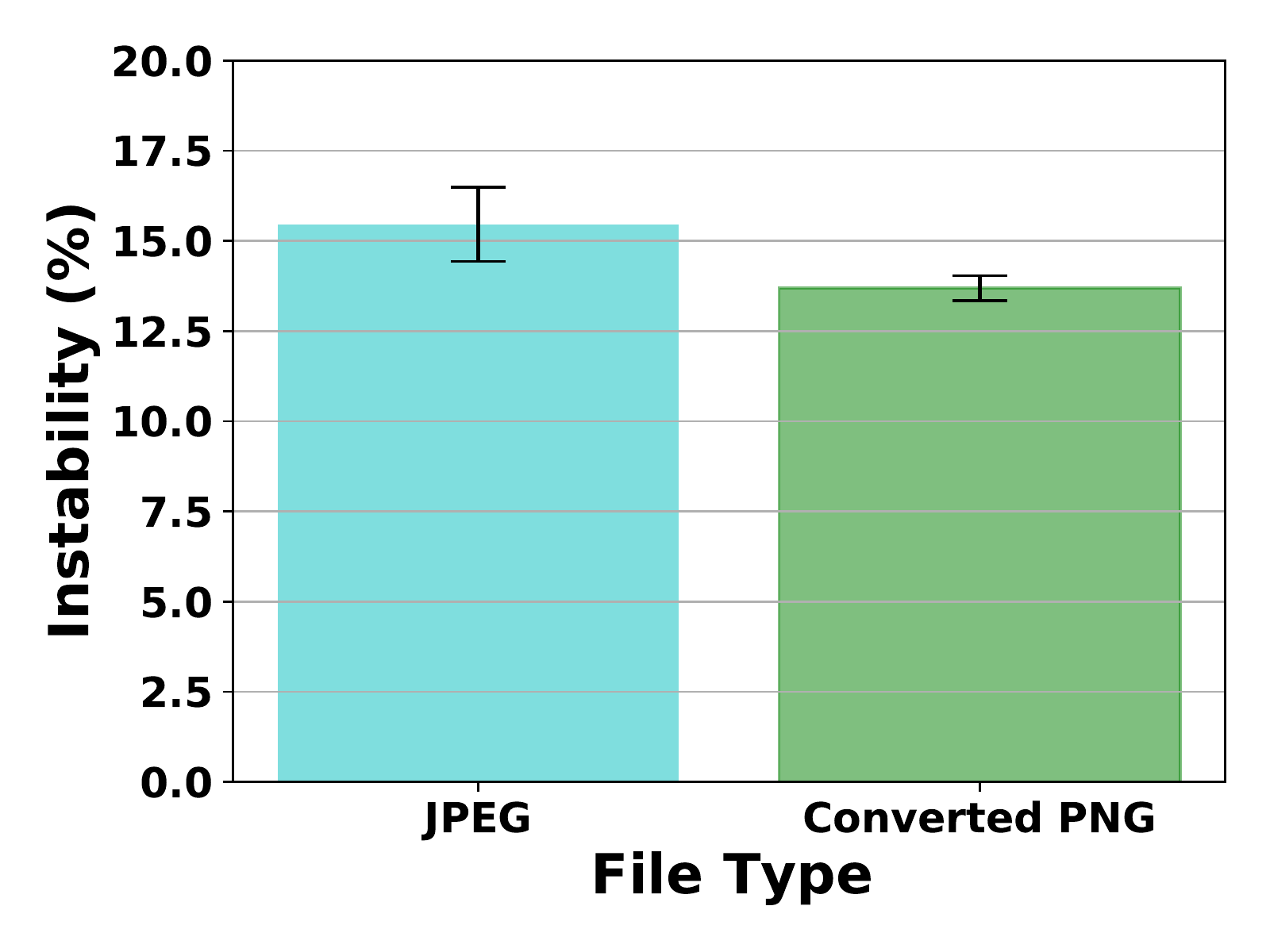}}
\subfigure[JPEG vs. raw converted images on Samsung and iPhone, broken by class.]{\label{fig:app-experiment-inconsistancy-b}%
\includegraphics[width=0.3\textwidth]{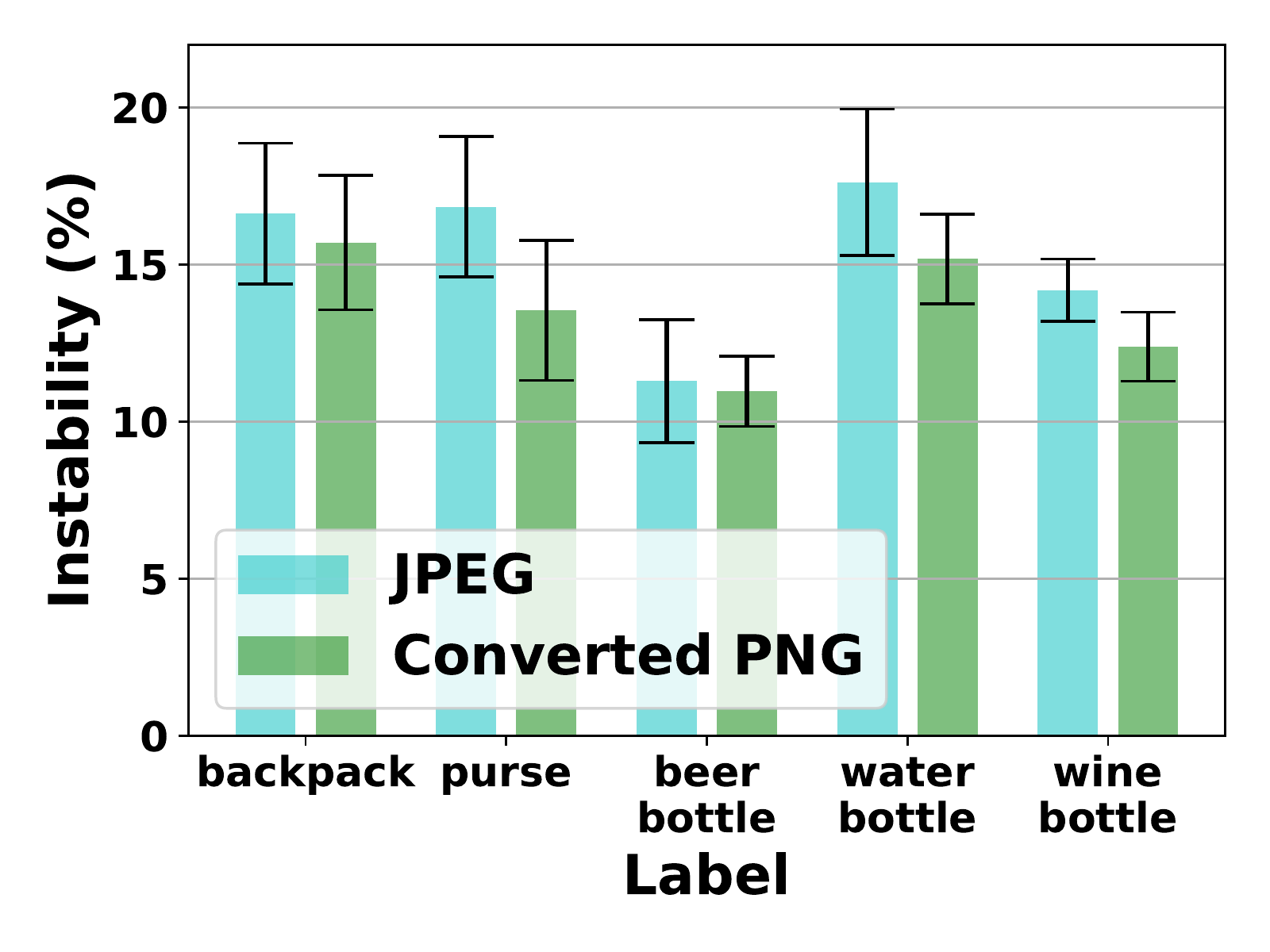}}
\subfigure[Accuracy of JPEG vs. raw converted images on Samsung and iPhone.]{\label{fig:app-experiment-inconsistancy-c}%
\includegraphics[width=0.3\textwidth]{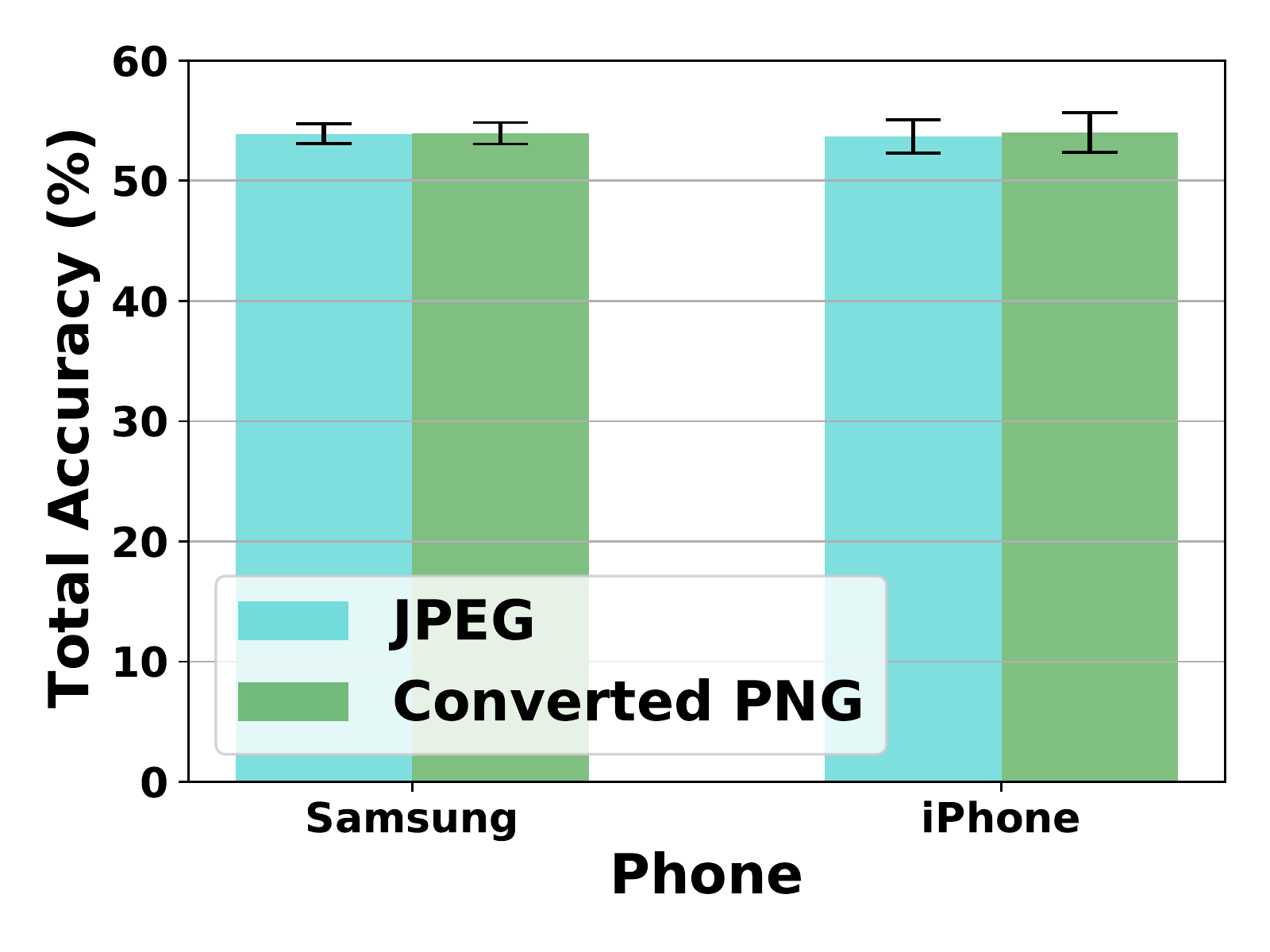}}
\vspace{-2.8pt}
\caption{Comparing JPEG and raw photos which where converted to PNG on iPhone and Samsung phones.}
\end{figure*}

In our final stability training experiment we similarly train on two images, one from Samsung and one from iPhone, but this time we limit the number of images we collect per class for iPhone. This version simulates how many images a developer would need to collect to adjust their model to a new phone. For example if we have a dataset from Samsung phones, it may be possible to augment it by collecting a limited amount of photos per class (\eg 10) from an iPhone to make the model more stable. We treat the number of images per class as a hyper parameter. This version is similar to the two image version of our model but for every image from Samsung, instead of supplying to the model the corresponding iPhone image, we pick one image from a small subset of iPhone images for the same class.

We trained a model for each version of noise generation with each version of stability loss on Samsung photos taken from the end-to-end experiment. For the versions of the training that require noisy input images we use the photos from the iPhone. For a base model we use a MobileNetV2 pre-trained on ImageNet. We test the result on photos taken both from Samsung and iPhone. We found our hyper parameters for the models using grid search. We compare all the stability trained model with regular fine-tuning of the base model on the Samsung phone without a stability loss. To evaluate the embedding distance loss we added one extra fully-connected layer to the base model. 

Table~\ref{tab:stability-results} presents the instability results across all noise generation schemes and both stability losses. We can see from the results that fine-tuning without stability training (denoted as ``No noise'') reduces instability the least. The results also demonstrate that including images from the iPhone for every image from the Samsung phone, and using embedding distance loss reduces instability the most. Yet, collecting many photos from each phone is not realistic.

Fortunately, augmenting the model with a modest number of photos per class from an iPhone (10 photos for each class), and using embedding distance loss resulted in fairly similar instability to collecting the entire dataset from an iPhone. We get 4.22\% instability with 10 photos per class, vs. 3.91\% with the entire dataset.

This solution though still requires calibration photos from each new phone we encounter. If that is not possible, using an image distortion noise with a relative entropy loss still provides a significantly lower instability (4.52\%) than the baseline, and requires no new data collection.

Figure~\ref{fig:pr-curves} contains the precision-recall graphs for all models trained under different noise generation schemes and stability losses. Interestingly, stability training has a small accuracy benefit as well as a benefit to instability. The two fine-tuning modes that augment the Samsung photos with iPhone photos provide the highest accuracy benefit.

\subsection{Using Raw Images in Inference\label{sec:using-raw}}

Our second approach is to test whether using raw images might reduce the level of instability, by comparing instability between JPEG and raw images. We repeated our end-to-end experiment but this time have each phone take both a compressed JPEG image and a raw DNG image, which is then converted to PNG. This limited the experiment to only the iPhone and Samsung phones, since they are the only phones capable of taking raw images. The conversion is done in a consistent manner for both phones using ImageMagick~\cite{imagemagick}, to eliminate any differences that may arise from using different ISPs between the phones.

Figure~\ref{fig:app-experiment-inconsistancy-a} and Figure~\ref{fig:app-experiment-inconsistancy-b} compare the instability between photos of the same object on both phones using MobileNetV2. The results demonstrate that using raw images does indeed slightly reduce the instability, both in the aggregate and in every class independently, but not by a significant amount. 


Figure~\ref{fig:app-experiment-inconsistancy-c} compares the accuracy of JPEG vs. raw converted images. This experiment interestingly shows that instability and accuracy are not necessarily correlated. While results on photos taken from both phones had similar accuracy, with iPhone photos producing only slightly better results, the instability was much higher than the accuracy difference. Additionally, using converted raw images did not lead to significant changes in accuracy. 

Utilizing raw images and a consistent conversion pipeline results in an average 11.5\% improvement in instability. Consistently throughout all experiments utilizing raw images outperformed using the phone's pipeline. 

However, utilizing raw images didn't eliminate instability completely. This implies, as described in \S\ref{sec:ISP}, that even in phones with raw image access, it is not always clear at what stage of the pipeline we get the raw image from. In addition, utilizing raw images requires hardware support from the devices. For instance in our end-to-end experiments only two of the five devices where able to take raw images.

\subsection{Simplifying the Classification Task}

\begin{figure}[h]
\centering
\subfigure[Top 3 prediction accuracy for end-to-end experiment.]{
\includegraphics[width=0.47\columnwidth]{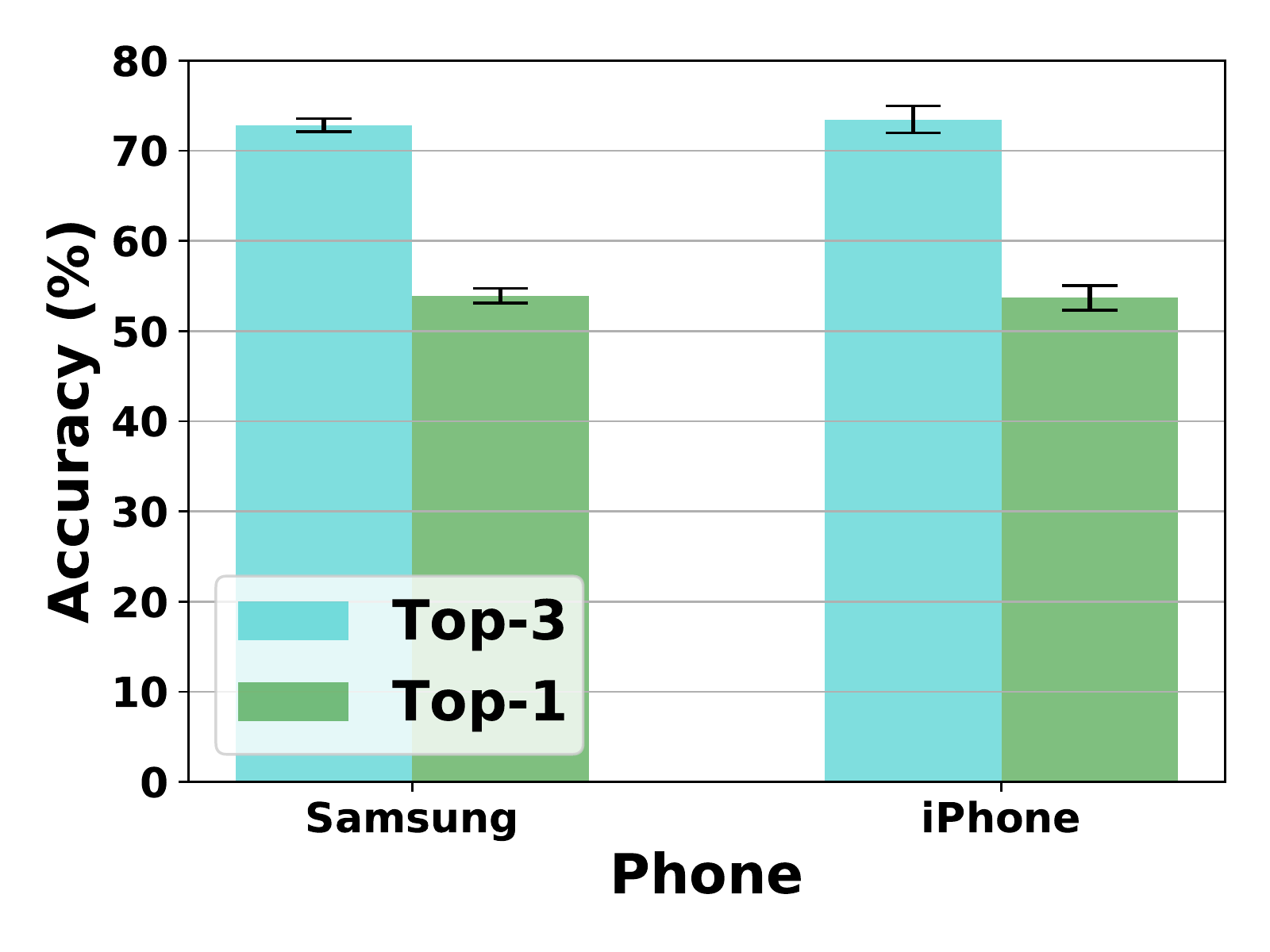}
\label{fig:top3-accuracy}}
\hfill
\subfigure[Top 3 prediction instability for end-to-end experiment.]{
\centering
\includegraphics[width=0.47\columnwidth]{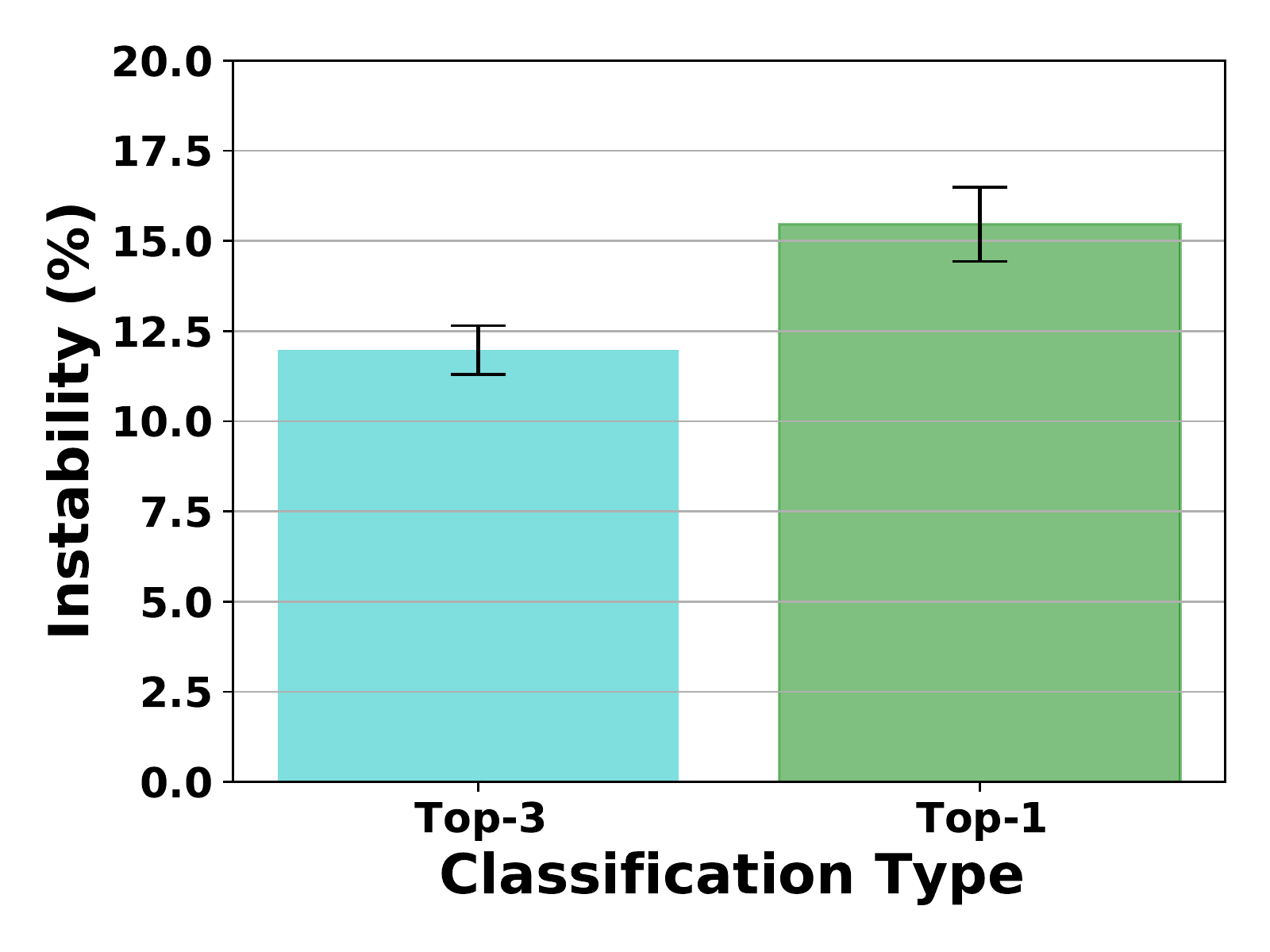} 
\label{fig:top3-inconsistancy}}
\vspace{-2.8pt}
\caption{Accuracy and instability of a model that displays the top three classifications compared to only the top one.}
\end{figure}

The last approach to reducing instability is to modify the prediction itself. For many models, such as recommendation systems or document search, it might be good enough for the correct prediction to be in one of the top $n$ predictions, rather than being the top result. To test this approach, using the same experimental setup we used before, we test the accuracy of having the correct classification appear in one of the top three results (Figure~\ref{fig:top3-accuracy}). Unsurprisingly we achieve a higher accuracy. Similarly, in Figure~\ref{fig:top3-inconsistancy} we can see that instability is also improved when we use top three results. Both accuracy and instability are improved by about $30\%$. Unfortunately, such task simplification is not viable for every application, and even for those where it is feasible, requiring the user to sift through additional possible classification results may degrade the user experience.

\section{Related Work}
Deploying deep learning on edge devices is a widely researched topic. Due to the limited resources of the edge devices, prior work focused on different techniques to make DNN models smaller and faster. MobileNet~\cite{MobileNet, mobilenetv2} and SqueezeNet~\cite{iandola2016squeezenet} are examples of models optimized to run on resource-limited devices. Another way to improve the size and performance of the models is by using techniques such as quantization~\cite{DBLP:journals/corr/HanMD15, BinarizedNN, Ternary-Weight-Networks, DBLP:conf/cvpr/JacobKCZTHAK18} and pruning~\cite{Optimal-Brain-Damage,DBLP:conf/nips/HassibiS92, DBLP:conf/nips/HanPTD15,DBLP:journals/corr/HanMD15}. However, these prior works do not consider the variability of running the model across different devices. 


Prior work from Facebook~\cite{FB-ML-inference-on-edge} presents the challenges of deploying machine learning algorithms on heterogeneous environments, such as mobile phones. They focus on the hardware challenges of accelerating ML on the edge, while maintaining performance, accuracy and high user experience. Their work is complementary to ours, as it mostly focuses on the latency and throughput of the model.

Past work examined the effects of image signal processors (ISPs) on deep learning accuracy. Buckler \etal~\cite{buckler2017reconfiguring} evaluated the effects of different stages of the ISP on accuracy and power usage, by creating a reversible ISP pipeline simulation tool. Using this analysis they created a low-power ISP for deep learning. Liu \etal~\cite{liu2015ultra} observe that perceived quality of an image is not correlated with accuracy of a CNN. Based on this insight they create a low-power ISP. Schwartz \etal~\cite{DeepISP} built an end-to-end ISP using a CNN by learning a mapping between raw sensor data and ISP processed images. To summarize, this set of works attempts to design a new ISP pipeline for deep learning. In contrast, our work treats ISPs as a black box, as they are largely outside the control of the developer and not consistent across devices.


There is a large body of work on deep learning robustness~\cite{nguyen2015deep, szegedy2013intriguing}. Akthar \etal survey the different papers in this area~\cite{survay}. The majority of work on DL robustness focuses on noise from adversarial examples, which are images specifically designed to produce incorrect results in models~\cite{kurakin2016adversarial, bastani2016measuring, cisse2017parseval}.

In contrast to work on adversarial robustness, DeepXplore~\cite{deepxplore} and DeepTest~\cite{deeptest} focus on how to train models to be more robust to environmental noise that may arise while driving, such as foggy road conditions. In our work we show that the device itself might add instability to the model. DeepCorrect~\cite{deepcorrect} focus on how to reduce Gaussian noise created during image acquisition. Meanwhile our work focuses on differences in noise between different devices. Dodge \etal~\cite{dodge2016understanding} study how models are affected by random Gaussian noise. Our fine-tuning model is inspired by prior work on stability training~\cite{StabilityTraining}, which tries to make models more robust to small perturbations in input data. We expand the work by adding a noise model designed to simulate differences between phones. We further show how to utilize stability training with minimal data collection.



\section{Conclusions}

This work examines the source of variation of model inference across different devices. We show that accuracy is a poor metric to account for this variation, and propose a new metric, instability, which measures the percentage of  nearly-identical inputs producing divergent outputs. We demonstrate that for classification, different compression formats and ISPs account for a significant source of instability. We also propose a technique to fine-tune models to reduce the instability due to different devices.

We believe further exploring other sources of instability is an important topic for future work. Other potential sources, which are beyond the scope of our paper, include variations in cameras and lenses, lighting and visibility conditions.

\balance
\bibliography{mlsys2021}
\bibliographystyle{mlsys2020}


\end{document}